\documentclass[10pt,twocolumn,letterpaper]{article}

\usepackage[pagenumbers]{cvpr} 

\definecolor{cvprblue}{rgb}{0.21,0.49,0.74}
\usepackage[pagebackref,breaklinks,colorlinks,allcolors=cvprblue]{hyperref}
\usepackage{adjustbox}

\usepackage{tocloft}
\def\paperID{1461} 
\def\confName{CVPR}
\def\confYear{2025}

\title{Incremental Object Keypoint Learning}

\author{Mingfu Liang$^1$ \quad Jiahuan Zhou$^{2,*}$ \quad Xu Zou$^3$ \quad Ying Wu$^1$  \\
$^1$ Northwestern University \quad $^2$ Wangxuan Institute of Computer Technology, Peking University \\ 
$^3$ Huazhong University of Science and Technology
}

\begin{document}
\maketitle
\begin{abstract}
Existing progress in object keypoint estimation primarily benefits from the conventional supervised learning paradigm based on numerous data labeled with pre-defined keypoints. However, these well-trained models can hardly detect the undefined new keypoints in test time, which largely hinders their feasibility for diverse downstream tasks. To handle this, various solutions are explored but still suffer from either limited generalizability or transferability. Therefore, in this paper, we explore a novel keypoint learning paradigm in that we only annotate new keypoints in the new data and incrementally train the model, without retaining any old data, called \textbf{I}ncremental object \textbf{K}eypoint \textbf{L}earning~(IKL). A two-stage learning scheme as a novel baseline tailored to IKL is developed. In the first \textit{Knowledge Association} stage, given the data labeled with only new keypoints, an auxiliary KA-Net is trained to automatically associate the old keypoints to these new ones based on their spatial and intrinsic anatomical relations. In the second \textit{Mutual Promotion} stage, based on a keypoint-oriented spatial distillation loss, we jointly leverage the auxiliary KA-Net and the old model for knowledge consolidation to mutually promote the estimation of all old and new keypoints. Owing to the investigation of the correlations between new and old keypoints, our proposed method can not just effectively mitigate the catastrophic forgetting of old keypoints, but may even further improve the estimation of the old ones and achieve a positive transfer beyond anti-forgetting. Such an observation has been solidly verified by extensive experiments on different keypoint datasets, where our method exhibits superiority in alleviating the forgetting issue and boosting performance while enjoying labeling efficiency even under the low-shot data regime. {\let\thefootnote\relax\footnote{{*: Corresponding author}}}
\end{abstract}    
\section{Introduction}
\label{Introduction}

\begin{figure}[t]
    \centering
    \includegraphics[width=0.8\linewidth]{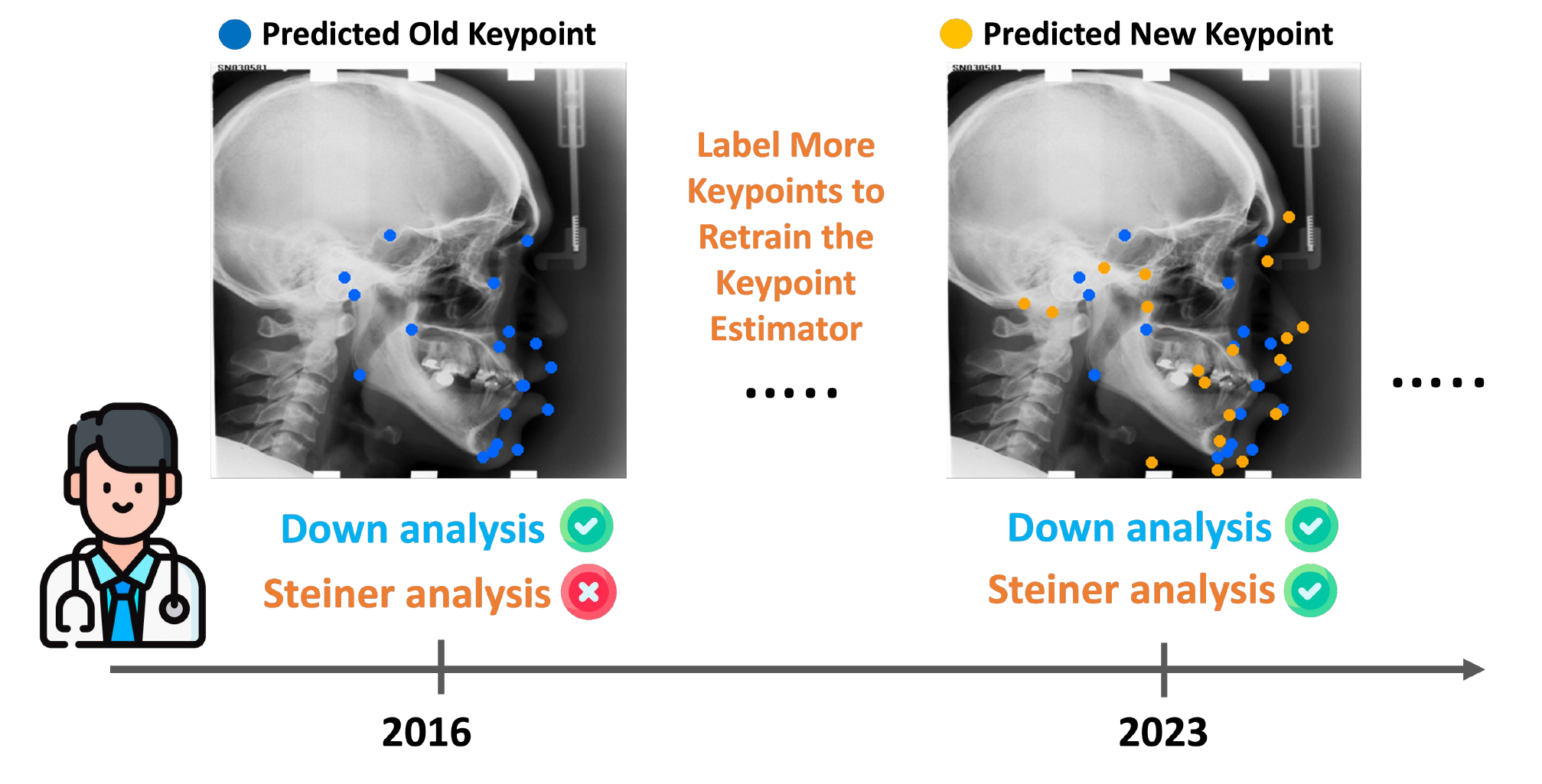}
    \vspace{-15pt}
    \caption{Medical analysis can frequently change and require new keypoints essentially~\cite{Cao2023-wf}, while the labeling is time-consuming.}
    \label{fig: examples of keypoints}
    \vspace{-10pt}
\end{figure}

\begin{figure*}[ht]
	\centering
	\includegraphics[width=1.0\linewidth]{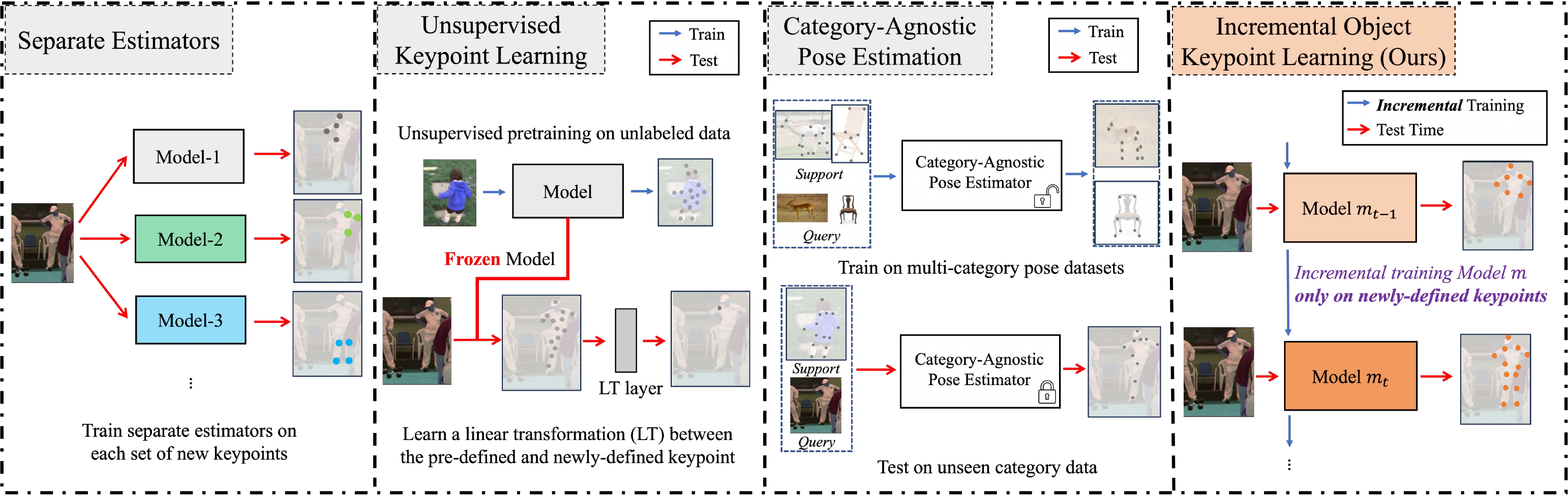}
        \vspace{-15pt}
	\caption{Separate Estimators need multiple estimators in test time. Unsupervised keypoint learning~(UKL) and category-agnostic pose estimation~(CAPE) exploit a fixed pretrained keypoint estimator on unseen new keypoints. While our Incremental object Keypoint Kearning~(IKL) continually updates the same model on the new data labeled \textbf{only on new keypoints without retaining any old training data}.}
    \label{framework}
    \vspace{-15pt}
\end{figure*}

As a fundamental task in computer vision, estimating the visual keypoint locations of an object serves as the indispensable primitive to support numerous down-stream tasks, e.g., object pose detection~\cite{jin2020whole,yang2021transpose,yu2021apk}, tracking~\cite{jiang2022avatarposer}, action recognition~\cite{carreira2017quo}, generation~\cite{siarohin2018deformable} and animation~\cite{siarohin2019animating,jiang2022avatarposer}, etc. Over a long period of time, \textbf{supervised keypoint learning (SKL)}~\cite{wang2020deep,geng2021bottom,li2021pose,li2021tokenpose} has significantly advanced the progress on keypoint estimation. By exploiting large-scale pure 2D keypoint datasets~\cite{andriluka14cvpr,lin2014microsoft} with extreme variance, SKL aims to train a deep neural network that can robustly detect a set of pre-defined keypoints in test images. 

However, the SKL models can not estimate newly-added undefined keypoints of an object, while new demands from downstream tasks will inevitably require new keypoints. For instance, until now, models trained by existing keypoint dataset~\cite{Wang2016-bn} can only detect 19 keypoints to support Downs analysis in the Cephalometric analysis, while they are infeasible for important analysis like Steiner analysis~\cite{Robert1982-nt}, as shown in Fig.~\ref{fig: examples of keypoints}. Such an issue motivated a recent MICCAI challenge~\cite{Cao2023-wf} on annotating much more keypoints to support increasing analysis. However, keypoint labeling is time-consuming \cite{durao2015cephalometric,liu2017active,moskvyak2021semisupervised}, especially when keypoints are defined sequentially. Another alternative solution is to train separate new keypoint estimator \cite{dai2007detector,karlinsky2012using} for each set of new keypoints as in Fig.~\ref{framework}. However, the number of estimators will increase linearly, and the separate training strategy will make each estimator easily overfit the new keypoints \cite{dai2007detector} and hardly capture the intrinsic relation between old and new keypoints, leading to sub-optimal performance.

To mitigate the above issues, \textbf{unsupervised keypoint learning (UKL)} \cite{zhang2018unsupervised,sanchez2019object,mallis2020unsupervised,he2022autolink,He_2023_CVPR} is proposed for new keypoint estimation. The models are firstly pretrained in an unsupervised manner to predict numbers of keypoints randomly, then a linear transformation is learned to transfer the existing keypoints to the new undefined ones. However, UKL is merely effective on objects with relatively rigid motion or video data with consistent object instances and local changes in appearance and motion. As a result, it is hard to perform UKL on large-scale pure 2D datasets~\cite{andriluka14cvpr,lin2014microsoft} due to their extreme inter-image differences in the background, appearance, and motions of distinct object instances. Recently, a supervised metric learning-based~\cite{snell2017prototypical} paradigm named \textbf{category-agnostic pose estimation~(CAPE)} \cite{xu2022pose} is proposed to learn a class-agnostic pose estimator that can estimate unseen keypoints in test time with a few labeled support samples. However, as the model is frozen after pretraining, CAPE is limited by its restricted generalization and transferability for newly-defined keypoints~(Tab.~\ref{comparison to few shot 2D methods}).

Therefore, in this paper, we propose to solve the new keypoint estimation problem creatively in an incremental learning~(IL) manner. As shown in Fig.~\ref{framework}, we propose an \textbf{Incremental object Keypoint Learning~(IKL)} paradigm to incrementally train a keypoint estimator on the \textbf{new data labeled only with new keypoints}. Even without retaining any old data, the IKL model should not forget the old keypoints catastrophically. Compared to the naive solutions that either label both new and old keypoints or learn separate estimators, our IKL only labels the new keypoints and maintains \emph{just one model in the lifetime}, which is computationally efficient. Moreover, different from the UKL and CAPE whose performances are largely limited by their restricted transferability depending on a fixed pretrained model, the IKL alleviates such an issue by continually training the model to sufficiently expand its knowledge on newly-defined keypoints and may \emph{even improve the old keypoint estimation} when learning more relevant new keypoints. Lastly, as most works in IL mainly study the classification task~\cite{delange2021continual,wang2024comprehensive}, to our best knowledge, IKL as an incremental keypoint regression task has rarely been studied before, which also provides novel insights for IL. 

However, IKL also poses a new challenge known as label non-co-occurrence (LNCO). As only new keypoints are labeled in the new data, the kinematic and anatomical constraints between old and new keypoints are not explicitly presented in the label space, making it hard for the model to capture such a physical prior during incremental training. Existing IL methods~(Tab.~\ref{tab: four dataset}) do not explicitly model such inter-keypoint relations in IKL, thus the old and new keypoint predictions can not mutually support each other and may readily bias toward the new ones, exacerbating the forgetting of the old keypoints~(Tab.~\ref{tab: four dataset}).

To tackle the challenge of IKL, we propose a novel two-stage learning scheme called \textbf{KAMP} as a new baseline tailored to IKL. 
In the first \textbf{K}nowledge \textbf{A}ssociation stage, we train an auxiliary KA-Net to associate old and new keypoints based on their physiological connection, represented by their spatial adjacency. 
Specifically, KA-Net learns to predict the selected old keypoint given the related new keypoints to acquire their intrinsic anatomical relevance. 
In the second \textbf{M}utual \textbf{P}romotion stage, we train the new model on all keypoints to mutually promote their estimation, where the old ones are updated by distilling from the KA-Net and the old model with a newly designed keypoint-oriented spatial distillation loss for better knowledge preservation. 
To verify our effectiveness, we simulate IKL based on four keypoints datasets on both medical and natural images, i.e., Cephalometric~\cite{Cao2023-wf}, Chest~\cite{Jaeger2014-zi}, MPII~\cite{andriluka14cvpr} and ATRW~\cite{li2020atrw}. 
Extensive results demonstrate that our KAMP can not just effectively alleviate the catastrophic forgetting of old keypoints, but even further boost their performance and thus outperform the existing exemplar-free IL methods by a significant margin~(Tab.~\ref{tab: four dataset}) and can also work for low-shot scenarios~(Tab.~\ref{comparison to few shot medical methods} and~\ref{comparison to few shot 2D methods}). 
Our analysis further reveals that IKL is complementary to learning paradigms like CAPE~(Tab.~\ref{comparison to few shot 2D methods}) and labeling-efficient for new keypoint estimation. 

To summarize, our contributions are three-fold: (1)~We establish a novel paradigm, Incremental object Keypoint Learning (IKL), to tackle the challenging demands of new keypoints estimation. (2)~We propose a two-stage exemplar-free IKL method to explicitly model and exploit the relation between old and new keypoints to help alleviate the label non-co-occurrence problem tailored to IKL, which may further boost the performance of old keypoints beyond anti-forgetting. (3)~As a proof-of-concept, we empirically show that IKL is practical and labeling-efficient to scale up a pretrained keypoint estimator on new keypoint estimation, which is much better than other alternative paradigms.

\section{Related Works}
\textbf{Keypoint Estimation~(KE).} Existing research of KE focuses on estimating a fixed number of keypoints for a specific category~\cite{yu2021apk,andriluka14cvpr,lin2014microsoft,sagonas2013300,jin2020whole,yin2022one,yao2021one}. 
Most works design new methods~\cite{zou2019learning,gu2021removing,li2021localization,li20212d,li2021human} or network architectures~\cite{tang2018deeply,tang2019does,yang2021transpose,luo2021rethinking,wang2020deep,li2021tokenpose} to improve the supervised learning over large-scale pure 2D keypoint datasets~\cite{lin2014microsoft,andriluka14cvpr}. 
Estimating dense keypoints~\cite{jin2020whole,zauss2021keypoint} of the human body has been proposed but requires extreme labeling cost. Recent works explore semi-supervised~\cite{honari2018improving,moskvyak2021semisupervised,wang2022pseudolabeled} or unsupervised learning~\cite{thewlis2017unsupervised,zhang2018unsupervised} to reduce the laborious labeling consumption, but the unsupervised methods still hardly achieve a precise estimation of semantically meaningful keypoints~\cite{mallis2020unsupervised}. 
Recently, Category-agnostic keypoint estimation (CAPE)~\cite{xu2022pose,Shi_2023_CVPR,Chen_2024_CVPR} is proposed to learn a class-agnostic model to detect keypoints of unseen categories by a few labeled support images without retraining the model. However, as the model is frozen after pretraining, CAPE is limited by restricted generalization and transferability due to the intra-class appearance variation, self-occlusion, and appearance ambiguity, as detailed in~\cite{xu2022pose}. Different from the above paradigm that pre-defined a set of keypoints or assumed the pretrained model was sufficient enough for different downstream tasks, our proposed IKL paradigm continuously updates the model only on the new keypoints to increasingly expand its knowledge, which is much more flexible and labeling-efficient.

\textbf{Incremental Learning~(IL).} Existing IL methods can be categorized as exemplar-free and exemplar-based ones~\cite{Aljundi2017CVPR,prabhu2020gdumb,kim2020imbalanced,hou2019learning,liang2022balancing,huang2025flat}, given whether old data can be retained. However, saving exemplars not just introduces increasing costs on memory, but also has privacy issues in real-world applications. For exemplar-free methods, existing approaches focus on adding regularization over parameters~\cite{kirkpatrick2017overcoming,zenke2017continual,lopez2017gradient,ahn2019uncertainty,yu2020semantic} or the network's output in different positions~\cite{dhar2019learning,zhao2020maintaining,Wu2021ICCV,li2017learning}. Regarding the IL settings, adding new classes~\cite{delange2021continual} or domains~\cite{lomonaco2017core50,volpi2021continual,liang2022balancing} have been studied popularly and mainly tested on classification benchmarks~\cite{deng2009imagenet,delange2021continual}. 
In contrast, keypoint estimation is a location regression task~\cite{gu2021removing,luo2021rethinking,li2021human}, and our proposed IKL aims to detect newly defined keypoints incrementally. The incremental animal pose estimation~\cite{nayak2021incremental}~(IAPE) is a related setting to us. However, IAPE assumes the label space of keypoints is \textbf{fixed} for each animal, and it only learns new animals' poses, i.e., only the input distribution is changed with new animal domains. While for our IKL, we instead consider \textbf{adding new keypoints} to the label space for an object category, and the old and new data distribution may be relatively correlated.

\textbf{Positive Transfer~(PT) in IL.} Most IL approaches focus on alleviating the catastrophic forgetting problem~\cite{delange2021continual}, while only a few works are dedicated to achieving positive transfer~(PT) in IL~\cite{lopez2017gradient,kao2021natural,lin2022beyond}, i.e., improving the old tasks while learning a new task. Those methods encourage PT by modifying the gradient updates based on the new and old tasks' correlations which are measured by saving the data~\cite{lopez2017gradient} from the old task or based on formal analyses~\cite{kao2021natural,lin2022beyond}. Differently, our KAMP achieves PT by explicitly \textbf{modeling the relation between old and new keypoints} based on their intrinsic anatomical relevance, and we \textbf{exploits the relation to design novel distillation to further boost the estimation of the old keypoint}, which is tailored to the IKL and keypoint estimation.
\section{Incremental Object Keypoint Learning}
\subsection{Problem Formulation}
We denote $t$=$0$ as the initial step and $t$=$1,...,T$ as the incremental steps. Then the training set for the $t$-th incremental step is $D^{\text{train}}_{t}$=$\left\{x_{t, j}^{\text {train }}, y_{t, j}^{\text {train }}\right\}_{j=1}^{N_t}$, where $x,y,N_t$ denote the inputs, the keypoint labels and the size of $D^{\text{train}}_{t}$. As we only label new keypoints in IKL, thus the newly introduced keypoints must be disjointed with the old ones, i.e., $y_t^{\text {train }} \cap\left(\cup_{i=0}^{t-1} y_i^{\text {train }}\right)=\varnothing$. Different from the classification task, the common practice for keypoint estimation is to regress a 2D heatmap for a keypoint, i.e., a Gaussian peak around the ground-truth keypoint location~\cite{wei2016convolutional}. Thus the model $m_{t}$ in IKL comprises the feature extractor $f_{t}$ and a stack of convolutional keypoint estimation heads $\{\mathrm{G}_{i}\}^{t}_{i=1}$ that parameterized by $\theta_{t}$ and $\{\phi_{i}\}^{t}_{i=1}$, respectively. Each estimation head ends with a convolutional layer to generate the 2D heatmap. $\mathrm{G}_{t}$ denotes the estimation head for the new keypoints in the $t$-th step. $f_{t}$ and $\{\mathrm{G}_{i}\}^{t-1}_{i=1}$ in model $m_{t}$ are initialized by the same weights in the old model $m_{t-1}$.

\begin{figure*}[t]
    \centering
    \includegraphics[width=0.85\linewidth]{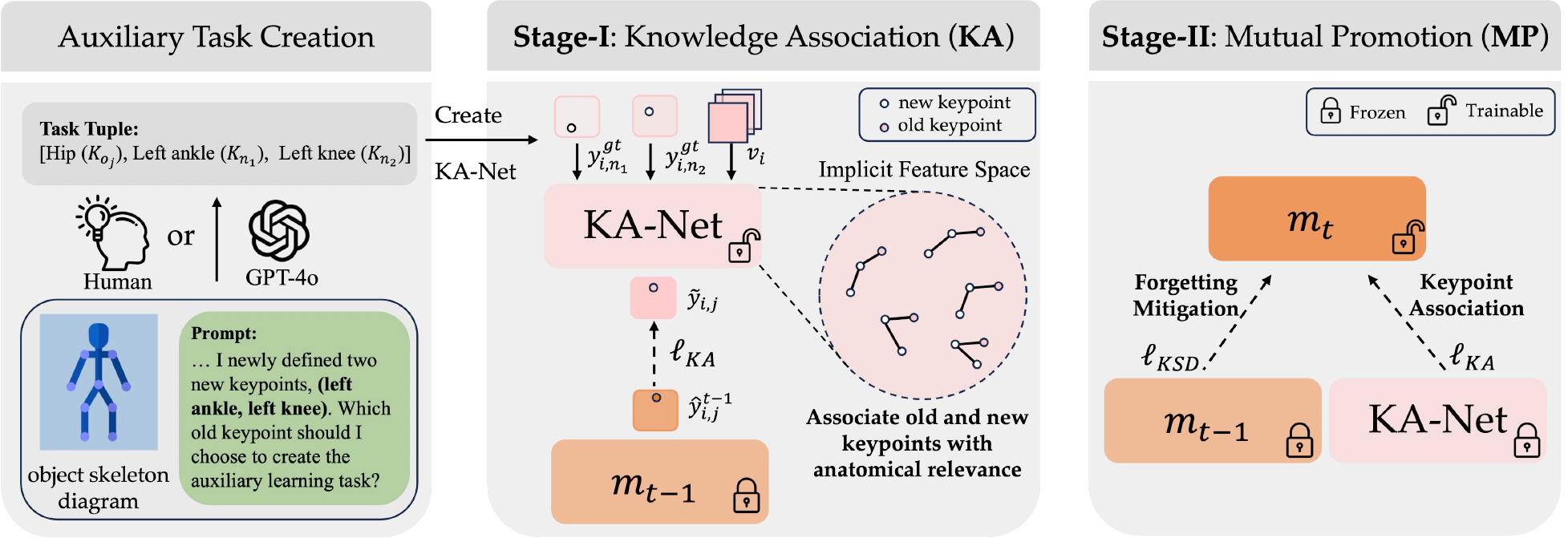}
    \caption{Overview of KAMP using the human body for illustration. In Stage-I, we learn an auxiliary KA-Net to associate the related old and new keypoints based on their local anatomical constraint. In Stage-II, we jointly leverage the old model and the KA-Net as an auxiliary teacher to consolidate all old keypoints' prediction and also learn the new keypoints simultaneously to achieve mutual promotions.} 
    \label{fig:illustration of case 2 network and training}
    \vspace{-15pt}
\end{figure*}

\subsection{KAMP: A Novel Baseline for IKL}
In this section, we present our two-stage method KAMP as a novel baseline tailored to IKL. We conjecture that distilling the relation between related old and new keypoints into the model may help it implicitly capture the intrinsic physical prior to alleviate the label non-co-occurrence issue. To do so, for each incremental step, in Stage-I, we design an auxiliary prediction task to associate the related old keypoint's with the new ones by the KA-Net. In Stage-II, the old model $m_{t-1}$ and KA-Net are both frozen to distill knowledge into the new model $m_{t}$ for all old keypoints, and the new keypoints are learned concurrently to achieve mutual promotion. Fig.~\ref{fig:illustration of case 2 network and training} provides a schematic view of KAMP.

\subsubsection{Stage-I: Knowledge Association~(KA)}
\label{Stage-1}
As mentioned in Sec.~\ref{Introduction}, the old and new keypoints in IKL are not jointly labeled, making it difficult for the model to learn relationships between keypoints, such as their structural and anatomical relevance, using only the labels of new keypoints.
To mitigate this issue, we propose to model the constraint among the spatially and anatomically related old and new keypoints as an implicit function in the current incremental step $t$.
Existing analysises~\cite{dai2007detector,tompson2014joint,chu2016structured} show that keypoint association can be well-modeled by the triangulation constraint.
Motivated by this, we define a triangulation constraint $F$ on the distribution of three related keypoints:
\vspace{-5pt}
\begin{equation}
\label{implicit function}
    F(P(K_{i}), P(K_{j}), P(K_{k})) =0,
\end{equation}
where $P(K_{i})$ denotes the distribution of the keypoint $i$, and the tuple may include one or two new keypoints. However, as there may be multiple valid solutions (constraints) that satisfy the implicit function $F$, it is challenging to derive an analytical solution for $F$ in IKL. 

Therefore, we consider an alternative formulation, e.g., $P(K_{\text{o}_{j}}) = F(P(K_{\text{n}_{1}}), P(K_{\text{n}_{2}}))$, which is a special case of the Eqn.~\ref{implicit function} that we condition the distribution of an old keypoint $P(K_{\text{o}_{j}})$ on its related new keypoints $P(K_{\text{n}_{1}})$ and $P(K_{\text{n}_{2}})$. 
We focus on two new keypoints since only they have ground-truth labels, and the ground-truth supervision can help reduce the uncertainty in learning the constraints. Moreover, in practice, new tasks typically require more than one new keypoint for support. 
Nonetheless, we verify in our Supplementary that our approach performs effectively even with a single new keypoint.
To enhance this constraint, we model the conditional probability between the related old and new keypoints as an auxiliary prediction task:
\vspace{-5pt}
\begin{equation}
    \label{task creation}
    P(K_{\text{o}_{j}} \mid K_{\text{n}_{1}}, K_{\text{n}_{2}}, v )=\text{KA}(P(K_{\text{n}_{1}}), P(K_{\text{n}_{2}}), v ),
\end{equation} 
where $\text{KA}$ stands for the \textbf{K}nowledge \textbf{A}ssociation Network~(KA-Net). The insight of this auxiliary task is inspired by the fact that spatially adjacent and visually correlated keypoints can mutually predict each other as shown in many existing works~\cite{tompson2014joint,chu2016structured,tang2018deeply}.

\textbf{KA-Net.} The input of the KA-Net is the ground-truth heatmap~($y^{gt}_{i,\text{n}_{1}}$ and $y^{gt}_{i, \text{n}_{2}}$) of new keypoints $K_{\text{n}_{1}}$ and $K_{\text{n}_{2}}$ of an image $i$, and the output is the predicted heatmap $\tilde{y}_{i, j}$ of the old keypoint $K_{\text{o}_{j}}$. 
$v_{i}$ denotes the holistic visual features $v$ of the image $i$ comprised of the intermediate features extracted from the frozen feature $f_{t-1}$ and are re-scaled to the same spatial size as $y^{gt}_{i,\text{n}_{1}}$ and $y^{gt}_{i, \text{n}_{2}}$. 
The visual features $v_{i}$ are incorporated alongside spatial information ($y^{gt}_{i,\text{n}_{1}}$ and $y^{gt}_{i, \text{n}_{2}}$) because, in a 2D image, spatial coordinates alone are insufficient to determine a keypoint's location due to uncertainty stemming from factors such as camera angle, object appearance, and motion. 
By incorporating visual features, we provide additional contextual information that helps account for this uncertainty in keypoint associations.

In the present paper, we explore KA-Net's simplest construction to minimize its training cost: for each new keypoint, we first perform the element-wise multiplication between its ground-truth heatmap~($y^{gt}_{i,\text{n}_{1}}$ and $y^{gt}_{i, \text{n}_{2}}$) and the holistic features $v_{i}$ to obtain the keypoint-oriented spatial features. 
Then we concatenate the spatial features and feed them forward over three convolutional layers accompanied by the Batch Normalization~(BN) and ReLU to predict the selected old keypoint $K_{\text{o}_{j}}$. 
Note that the trained KA-Net will only be used to distill the keypoint association in the Stage-II and will \textbf{not} be used in test time after IKL.

\textbf{Training of KA-Net.} We train KA-Net by the new data $D^{\text{train}}_{t}$. As $D^{\text{train}}_{t}$ does not have ground truth labels for old keypoints, we use the pseudo-labels predicted by the old model $m_{t-1}$ to supervise KA-Net for keypoint regression:
\vspace{-5pt}
\begin{equation}
    \ell_{KA}= \frac{1}{N_t} \sum\nolimits_i^{N t} \sum\nolimits_{j \in \mathcal{K}_{KA}}\left\|\hat{y}_{i, j}^{t-1}-\tilde{y}_{i, j}\right\|_2^2,
\end{equation}
where $\mathcal{K}_{KA}$ denotes \textbf{the selected old keypoints} used for the auxiliary prediction task, $\hat{y}_{i, j}^{t-1}$ denotes the prediction of the $j$-th keypoint by $m_{t-1}$ given image $i$, and $\tilde{y}_{i, j}$ denotes the prediction by the KA-Net. 
We will show in Sec.~\ref{experiment} that our KAMP can effectively reduce the forgetting of old keypoints and even improve them, which largely reduce the accumulative error to use pseudo-labels for KA-Net.

\textbf{Auxiliary Task Creation.} To select the old keypoint for training KA-Net, in each incremental step, we first locate all the old and new keypoints on a general object anatomy diagram by their semantic definition. 
Then we measure their relative proximity by simple distance calculations~(e.g., Euclidean distance) to identify two new keypoints ($K_{\text{n}_{1}}$ and $K_{\text{n}_{2}}$) that are close to an old one $K_{\text{o}_{j}}$.
This process can also be automated by multi-modality large language models like GPT-4o~\cite{GPT-4o,xing2023toa,liang2024aide} as shown in Fig.~\ref{fig:illustration of case 2 network and training}.
If we identify several tuples of old and new keypoints satisfied the requirement, we randomly choose one tuple to create the auxiliary prediction task to avoid training too many KA-Net.
For all results of our KAMP in Section~\ref{experiment}, we show that even by creating only one auxiliary task in each step, we can still bring sufficient improvement on all keypoints. 
Note that task creation does not need any training, and we only need to perform once before each step, which is highly efficient. 
Such a design provides an interpretable way to incorporate the physical knowledge of the object to guide the IKL.

\subsubsection{Stage-II: Mutual Promotion~(MP)}
In Stage-II, we jointly optimize all the new and old keypoints on the new model $m_{t}$ by the loss $l_{M P}$:
\vspace{-5pt}
\begin{equation}
    l_{M P}=\ell_{G T}+\alpha\left(\ell_{KSD}+\ell_{KA}\right),
\end{equation}
where $\ell_{G T}$ denotes the L2 loss between the ground truth of the new keypoints and their predictions by the new model to acquire the knowledge of new keypoints. $\alpha$ is a hyperparameter to balance the new keypoints acquisition and old keypoints forgetting. 
For $\ell_{KA}$, we use the frozen KA-Net as an auxiliary teacher to supervised the old keypoint selected in Stage-I to distill the keypoint association knowledge. 
Since $\ell_{KA}$ is only applied to the selected old keypoint for knowledge transfer instead of mitigating the forgetting, we further consolidate the knowledge of all old keypoints by the loss $\ell_{KSD}$ using the frozen old model $m_{t-1}$. 
As the distribution of new and old data in IKL do not change dramatically and may even have a strong correlation, LWF~\cite{li2017learning} can be a baseline for $\ell_{KSD}$ in such a scenario as analyzed in \cite{delange2021continual}. 
In general, $\ell_{KSD}$ represents the negative log-likelihood $ \frac{1}{N_t}\sum_i^{N_{t}}s\left(\hat{y}_{i}^{t-1}\right) \cdot \log s\left(\hat{y}_{i}^t\right)$ between the predictions of all old keypoints by the new model, $\hat{y}_{i}^{t-1}$$\in$$\mathbb{R}^{C \times H \times W}$,  and predictions by the old model, $\hat{y}_{i}^t$$\in$$\mathbb{R}^{C \times H \times W}$. 
$s(\cdot)$ denotes the Softmax operator. $C$ denotes the numbers of old keypoints learned before step $t$, $H$ and $W$ are the height and width of each keypoint's heatmap. 

However, since LWF and its variants all focus on the classification task, they perform the Softmax across different classes by default, i.e., over the $C$ dimension, to obtain the normalized class prediction score. For IKL, this means normalizing across all old keypoints' predictions, which can not explicitly regularize the discrepancy of each old keypoint between the old and new model. To better preserve keypoint-specific knowledge, we adopt a spatial softmax operation over height~($H$) and width~($W$) dimensions of the keypoint prediction, $s^{\text{d}}_{\text{sp}}(\cdot), \text{d}\in{\{H,W\}}$, and combine them to encourage spatial-oriented knowledge distillation. 
We discuss in our Supplementary the difference between each softmax alternative. 
Thus, our $\ell_{KSD}$ is defined as
\vspace{-5pt}
\begin{equation}
   \frac{1}{N_t}\sum\nolimits_i^{N t}\sum\nolimits_j^{C}\sum\nolimits_{\text{d}}-s^{\text{d}}_{\text{sp}}\left(\hat{y}_{i,j}^{t-1}\right) \cdot \log s^{\text{d}}_{\text{sp}}\left(\hat{y}_{i,j}^t\right).
\end{equation}
\section{Experiment}

\begin{table*}[htbp]
\vspace{-10pt}
\scriptsize
  \centering
  \vspace{-10pt}
    \begin{tabular}{l|ccc|ccc|ccc|ccc}
    \toprule
          & \multicolumn{3}{c|}{\textbf{Split Chest }} & \multicolumn{3}{c|}{\textbf{Split Head-2023}} & \multicolumn{3}{c|}{\textbf{Split MPII}} & \multicolumn{3}{c}{\textbf{Split ATRW}} \\
\cmidrule{2-13}    \textbf{Method} & $\text{A-MRE}_{1}$~$\downarrow$   & $\text{AT}_{1}$~$\uparrow$    & $\text{MT}_{1}$~$\uparrow$   & $\text{A-MRE}_{4}$~$\downarrow$   & $\text{AT}_{4}$~$\uparrow$    & $\text{MT}_{4}$~$\uparrow$    & $\text{AAA}_{4}$~$\uparrow$   & $\text{AT}_{4}$~$\uparrow$    & $\text{MT}_{4}$~$\uparrow$    & $\text{AAA}_{3}$~$\uparrow$   & $\text{AT}_{3}$~$\uparrow$    & $\text{MT}_{3}$~$\uparrow$ \\
    \midrule
    Joint Training & 5.43  & -     & -     & 2.12  & -     & -     & 88.50 & -     & -     & 94.69 & -     & - \\
    Finetune & 43.1  & -     & -     & 51.3  & -     & -     & 37.41 & -     & -     & 13.24 & -     & - \\
    \midrule
    EWC~\cite{kirkpatrick2017overcoming}   & 13.28 & -8.23 & -3.67 & 10.97 & -6.37 & -4.76 & 38.64 & -51.84 & -12.21 & 14.38 & -59.75 & -2.08 \\
    RW~\cite{chaudhry2018riemannian}    & 9.48  & -7.12 & -4.15 & 6.49  & -4.23 & -2.88 & 38.47 & -18.83 & -7.13 & 84.15 & -10.87 & 0.00 \\
    MAS~\cite{aljundi2018memory}    & 7.36  & -1.86 & -0.19 & 5.31  & -2.15 & -1.33 & 67.03 & -7.56 & 0.34  & 85.68 & -5.80 & -1.13 \\
    LWF~\cite{li2017learning}   & 6.35  & -1.34 & 0.18  & 4.31  & -1.26 & 0.57  & 75.75 & -3.86 & 0.41  & 87.31 & -5.10 & -0.64 \\
    AFEC~\cite{wang2021afec}   & 8.04  & -2.67 & 0.15  & 5.77  & -3.45 & -1.46 & 37.24 & -22.85 & -15.42 & 33.03 & -40.25 & -8.02 \\
    CPR~\cite{cha2021cpr}   & 6.17  & -0.87 & 0.29  & 3.71  & -1.18 & 0.16  & 75.52 & -3.24 & 0.75  & 89.34 & -2.76 & 4.49 \\
    SFD~\cite{douillard2021plop} & 7.68 & -0.54 & 0.13 & 4.76 & -0.43 & 0.02 & 71.49 & -0.93 & 0.21 & 86.11 & -1.13 & 0.41 \\
    WF~\cite{xiao2023endpoints} & 7.31 & -0.31 & 0.16 & 4.58 & 0.03 & 0.11 & 72.87 & -0.46 & 0.38 & 86.69 & -0.97 & 0.62 \\
    GBD~\cite{dong2023heterogeneous} & 6.42 & 0.06 & 0.21 & 4.34 & 0.12 & 0.47 & 75.62 & -0.18 & 0.35 & 87.42 & -0.89 & 0.65 \\
    \midrule
    KAMP (Ours)  & \textbf{5.67} & \textbf{0.29} & \textbf{0.62} & \textbf{2.32} & \textbf{0.41} & \textbf{0.84} & \textbf{79.93} & \textbf{1.80} & \textbf{4.23} & \textbf{93.16} & \textbf{-0.84} & \textbf{5.13} \\
    \bottomrule
    \end{tabular}%
  \caption{Result of 4 datasets after 2-Step, 5-Step, 5-Step and 4-Step IKL for Chest, Head-2023, MPII, and ATRW respectively. All comparison methods are started from the same Step-0 trained model. $\text{A-MRE}$: smaller the better; $\text{AAA}$/$\text{AT}$/$\text{MT}$: larger the better.}
  \label{tab: four dataset}%
  \vspace{-10pt}
\end{table*}%
\label{experiment}
\textbf{Datasets.} As the IKL has rarely been studied before, there is no specific dataset or benchmark for IKL. Motivated by the real-world application of IKL in medical analysis, we leverage the large-scale Cephalometric keypoint dataset~\cite{Cao2023-wf} proposed in the 2023 MICCAI Challenge and also the commonly benchmarked Chest dataset~\cite{Jaeger2014-zi} to further create the IKL protocols to validate our proposed method. Note that the Cephalometric keypoint dataset~\cite{Cao2023-wf} differs from the head dataset~\cite{Wang2016-bn} used in previous works~\cite{yao2021one,yin2022one} commonly, as \cite{Cao2023-wf} has much larger number of keypoints and re-collected from different hospitals to enlarge the variance of images, makeing it much more challenging than \cite{Wang2016-bn}. Thus we term this dataset~\cite{Cao2023-wf} as Head-2023 for clarity. Note that both Head-2023 and Chest datasets have less than or equal to 400 images in total. Moreover, though there is not a large need for IKL on human and animal, the widely used human and animal keypoint datasets, i.e., MPII~\cite{andriluka14cvpr} and ATRW~\cite{li2020atrw}, have extreme variations and non-rigid motions. Thus we also choose them for experiments to show the generality of our KAMP under different domains and challenging scenarios. We detail full dataset statistics in our Supplementary. 

\textbf{Compared Methods.} For keypoint estimation, common network structures for classification like ResNet~\cite{he2016deep} need specific modifications, e.g., adding the deconvolution layers and convolution layers to generate the keypoint location heatmap. 
These adjustments make many IL methods inapplicable for IKL, e.g., prototype-based methods~\cite{zhu2021class,zhu2021prototype} like PASS~\cite{zhu2021prototype} that leverages the feature mean prototype before the linear classification layer. 
Furthermore, the IKL requires that the old data can not be retained, and it is hard to identify specific methods for IKL. 
Thus we choose several general and representative exemplar-free methods that can adapt to different IL settings based on their methodologies, including EWC~\cite{kirkpatrick2017overcoming}, LWF~\cite{li2017learning}, MAS~\cite{wu2018memory}, RW~\cite{chaudhry2018riemannian}, AFEC~\cite{wang2021afec}, CPR~\cite{cha2021cpr} for comparison. 
They are regularization-based methods and can be easily applied to the IKL without trial and error. 
We further adapt methods from Continual Semantic Segmentation~(CSS) and Class Incremental Learning~(CIL) to ISL for more comparisons, i.e., spatial feature distillation~(SFD)~\cite{douillard2021plop}, weight fusion~(WF)~\cite{xiao2023endpoints}, and gradient balanced distillation~(GBD)~\cite{dong2023heterogeneous}. 
We also report the native baseline, i.e., directly finetune the model during IKL, and the upper bound that trains the model~(e.g., HRNet~\cite{sun2019deep}) with all the data, where we denote the former as ``Finetune'' and the latter as ``Joint Training''. 

\textbf{Evaluation Metrics.} To assess the keypoint regression task, we employ the widely used mean radial error (MRE) for Head-2023 and Chest datasets as in~\cite{yao2021one,yin2022one}, and Probability of Correct Keypoint (PCK)~\cite{yang2021transpose,moskvyak2021semisupervised,wang2022pseudolabeled,Shi_2023_CVPR} for MPII and ATRW and use their defaulted $\sigma$ as in \cite{sun2019deep,moskvyak2021semisupervised} to compute PCK. To measure the performance of incremental learning (IL), we use Average Accuracy (AAA), calculating the accuracy~(\%) across all keypoints post-incremental step $i$ under PCK, and extend this approach to MRE, denoting it as A-MRE. Additionally, we introduce two metrics for knowledge transfer in IL: (1) \textbf{Average Transfer (AT)}, also known as backward transfer~\cite{lopez2017gradient}, which averages the performance improvement of keypoints over all previous steps after step $i$, and (2) \textbf{Maximal Transfer (MT)}, measuring the largest performance improvement in any old keypoint post-step $i$. Notably, when calculating AT and MT with MRE, we invert the sign of the change in error. This adjustment ensures consistency, since a decrease in MRE signifies improvement, in contrast to the direct correlation of increased accuracy with improvement in the PCK metric. Detailed explanations for the calculation of each metric are provided in our supplementary materials.

\textbf{Experimental Design.} We randomly split the keypoints of Chest, Head-2023, MPII, and ATRW into different incremental steps to create the Spilt Head-2023, Chest, MPII and ATRW protocols respectively. As the Chest dataset only has 6 keypoints in total, we split them into two groups where the first group has 4 keypoints with 150 training images, and the second group for IKL has 2 keypoints with only 50 training images. For Split Head-2023, we split the 38 keypoints into 5 groups, where the first group has 19 keypoints with the same definition as \cite{Wang2016-bn} with 100 training images, and we split the rest of keypoint into 4 groups randomly to simulate four incremental steps with only 50 training images for each step. Similarly, the 16 MPII keypoints and 15 ATRW keypoints were split into 5 and 4 groups, respectively, with an equal distribution of training images per step. We repeat five times under different orders of keypoint groups and report the mean value in the main paper. The standard deviation~(std), details about the keypoints group, and results of other experimental setups are in our Supplementary.

\textbf{Implementation Details.} We use the HRNet~\cite{sun2019deep} as the backbone for all methods as it is widely compared in keypoint estimation~\cite{moskvyak2021semisupervised,li2021tokenpose,li2021pose,geng2021bottom,wang2022pseudolabeled,yin2022one}. For a fair comparison, we initialize all methods with the same initial step~(Step-0) model for incremental learning. As the first benchmark for the IKL, we use the Continual Hyperparameter Framework~(CHF)~\cite{delange2021continual,liang2022balancing} to identify the training parameters like training epoch~(100), initial learning rate~(2e-3 for Split Chest and Head-2023, 1e-2 for Split MPII, 1e-3 for Split-ATRW), momentum~(0.9), weight decay~(1e-4), and also the hyperparameters of each compared methods, all included in our Supplementary. For our KAMP, the hyperparameter $\alpha$ is set as 1e2 for the Split Head-2023 and Split-MPII and 1e4 for the Split Chest and Split-ATRW in all experiments. The analysis of the $\alpha$ and other implementation details are included in our Supplementary.

\subsection{Comparison with SOTA for IKL}
\begin{figure*}[t]
	\centering
	\includegraphics[width=1.0\linewidth]{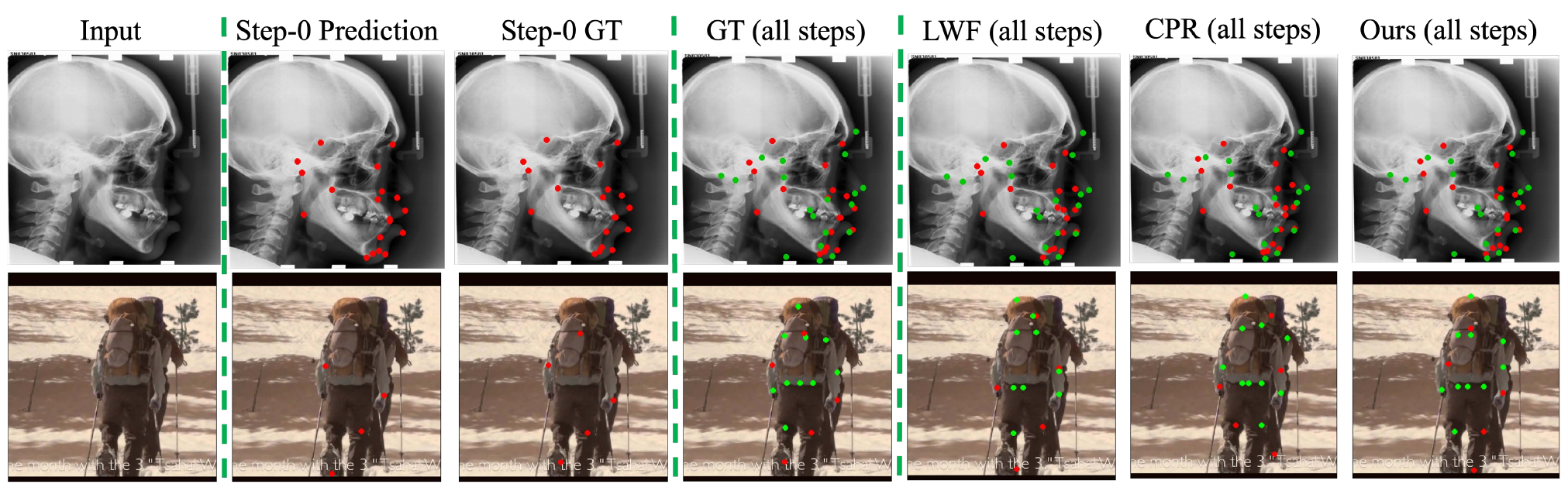}
        \vspace{-15pt}
	\caption{Qualitative results on Split Head-2023 and MPII. All methods start from the same Step-0 model, whose prediction is in the second column. GT: ground truth. The {\color{red}{red}} circles denote the keypoints learned in Step 0, and the {\color{green}{green}} ones denote all the new keypoints learned in later incremental steps. We observe that after IKL, the compared methods~(LWF and CPR) may acquire the new keypoints as ours, but they have obvious miss-detection and wrong estimation~(e.g., out of the body). While our method can consistently associate the new and old keypoints and achieve structurally accurate keypoint predictions. More results are included in our Supplementary.} 
	\label{fig: qualitative result}
        \vspace{-15pt}
\end{figure*}

As shown in Tab.~\ref{tab: four dataset}, our KAMP consistently improves accuracy and reduces error across all old and new keypoints after each incremental step, achieving the highest performance compared to all other methods on all datasets.
For example, KAMP outperforms the second-best method by \textbf{1.39\%} in $\text{A-MRE}_{4}$ after four incremental steps on Split Head-2023, and by \textbf{4.18\%} in $\text{AAA}_{4}$ on Split MPII. 
Moreover, KAMP is the only method to consistently yield the highest average transfer ($\text{AT}$) and maximal transfer ($\text{MT}$) scores across all datasets. 
This result highlights that our two-stage learning scheme better facilitates knowledge transfer, thereby enhancing accuracy on old keypoints while learning new ones. 
When $\text{AT}$ is positive, this indicates \emph{no forgetting on average}, underscoring the effectiveness of using pseudo-labels from the old model to train KA-Net, as described in Sec.~\ref{Stage-1}.
In Split-ATRW, the $\text{AT}$ is negative for all methods, indicating that forgetting of old keypoints outweighs knowledge transfer.
However, KAMP still achieves the highest $\text{AT}$ among the methods.  
Notably, the absolute value of KAMP's $\text{AT}$ is nearly zero, suggesting minimal negative transfer.
Additionally, by examining the maximal transfer metric, we observe a significant positive transfer in specific cases, such as \textbf{5.13\%} for $\text{MT}_{3}$, indicating that some old keypoints experience a substantial positive transfer in each incremental step.
This positive transfer helps offset the forgetting of other old keypoints, resulting in a minimal negative average transfer. We further validate this effect by reporting per-keypoint performance in our Supplementary.
Qualitative results in Fig.~\ref{fig: qualitative result} further verify our superiority in achieving more structurally correct prediction than competitive methods. 

Lastly, we observe that methods adapted from CSS and CIL, i.e., SFD, WF, and GBD, achieve strong performance on $\text{AT}$ and $\text{MT}$ metrics compared to competitive methods like LWF and CPR, but perform poorly on overall metrics like $\text{A-MRE}$ and $\text{AAA}$. 
This is because these methods prioritize reducing forgetting, resulting in an overly rigid model that struggles to effectively incorporate new keypoints in IKL.
This observation highlights a crucial limitation of existing CSS and CIL methods that they focus heavily on regularizing forgetting but fail to balance this with knowledge acquisition for new keypoints, which is especially problematic in our newly defined IKL setting. 
Consequently, this emphasizes the need to study IKL as a distinct incremental learning scenario in this paper, as traditional incremental learning methods are too general to excel here. 
\textbf{Our proposed novel baseline, KAMP, effectively addresses this gap for IKL, offering a well-balanced approach that goes beyond anti-forgetting to deliver robust performance across both accuracy and transfer metrics.}

\subsection{Ablation Study}
To verify the effectiveness of our method, we further analyze how each component may influence the result. Since our method contains two-stage training and we consider a keypoint-oriented spatial distillation loss~($\ell_{KSD}$), thus we compared our method with three alternatives: (1) the competitive baseline, i.e., the LWF, which performs the Softmax operation across old keypoints. (2) only use our $\ell_{KSD}$ to train the model without the KA-Net. 
(3) Constructing the KA-Net \textbf{randomly} without using physical knowledge. 
Results are shown in Tab.~\ref{ablation}, and we can observe that:~(1) our adapted $\ell_{KSD}$ can effectively outperform the LWF with 1.18\%, which demonstrates the essence of our keypoint-oriented adaptation.~(2) With proper auxiliary task creation based on physical prior, we can achieve more positive average transfer than other counterparts and finally achieve the largest improvement over all the keypoints on $\text{AAA}_{4}$. 
\begin{table}

\small
\centering
\begin{tabular}{cccc}
    \toprule
    \textbf{Method} & $\text{AAA}_{4}$~$\uparrow$     & $\text{AT}_{4}$~$\uparrow$     & $\text{MT}_{4}$~$\uparrow$  \\
    \midrule
    LWF~\cite{li2017learning}   & 75.75 & -3.86 & 0.41 \\
    KAMP~(only $\ell_{KSD}$) & 76.93 & -2.24 & 0.65 \\
    KAMP~(Random KA-Net) & 77.13 & -0.48 & 1.24 \\
    \midrule
    KAMP~(Ours)  & \textbf{79.93} & \textbf{1.80} & \textbf{4.23} \\
    \bottomrule
    \end{tabular}%
    \caption{Ablation Study on Split MPII.}
    \vspace{-15pt}
  \label{ablation}%
\end{table} 
This shows that our proposed two-stage learning scheme for IKL can not just provide better knowledge consolidation on the old keypoints than the competitive baseline, but the auxiliary prediction task can also bring large improvement to the old keypoints, showing that our KAMP will be a novel and strong baseline for the IKL paradigm.
More ablation studies like using backbones other than HRNet and more datasets are in our Supplementary.

\subsection{Compare IKL to other low-data paradigms}
Now we provide more insights into our IKL by comparing it with other learning paradigms with low-shot data. Discussion of limitations are in our Supplementary.  

As mentioned before, existing learning paradigms like UKL~\cite{he2022autolink,He_2023_CVPR,yao2021one} and CAPE~\cite{xu2022pose,Shi_2023_CVPR,Chen_2024_CVPR} leverage large-scale self-supervised and multi-dataset training to enable novel keypoint detection with a few labeled samples in test time. 
To fairly compare with UKL and CAPE, we extend KAMP to low-data regime following EGT~\cite{yao2021one}, where we also pretrain an auxiliary self-supervised model at Step-0 to provide pseudo-labels for old keypoints during IKL, and the details are in the Supplementary. 
For medical datasets like Head-2023, we compare to the SOTA one-shot methods CC2D~\cite{yao2021one} and EGT~\cite{yin2022one}. For 2D datasets like MPII, we compare to the SOTA Autolink~\cite{he2022autolink} and MetaPoint+~\cite{Chen_2024_CVPR} with official implementation available. Note that methods like \cite{zhou2018starmap} require extra expensive 3D CAD models to identify keypoint proposals for novel keypoint detection and are inferior to the SOTA~\cite{Shi_2023_CVPR,Chen_2024_CVPR}, thus we only compare to the MetaPoint+~\cite{Chen_2024_CVPR}. For KAMP, we use the low-data adaptation for 1, 5, and 10 shot and do not use it for 50-shot since 50-shot is large enough for us to bypass the adaptation. Full experimental details are in our Supplementary.

\begin{table}[htbp]
 \small
  \centering
    \begin{tabular}{l|cccc}
    \toprule
    \textbf{Method/(MRE~$\downarrow$)} & 1-shot & 5-shot & 10-shot & 50-shot \\
    \midrule
    CC2D~\cite{yao2021one}  & 5.14  & 4.83  & 4.08  & 3.47 \\
    EGT~\cite{yin2022one}   & 5.01  & 4.58  & 3.87  & 3.21 \\
    \midrule
    KAMP (Ours) & \textbf{4.35}  & \textbf{3.70}  & \textbf{3.03}  & \textbf{2.32} \\
    \bottomrule
    \end{tabular}%
    \caption{Compare to low-shot methods on Split Head-2023.}
  \label{comparison to few shot medical methods}%
  \vspace{-10pt}
\end{table}%

From Tab.~\ref{comparison to few shot medical methods} we observe that even on the extreme 1-shot setting, KAMP still outperforms the CC2D and EGT, while all the compared methods do not have good performances. 
This is because the Head-2023 dataset is collected from multiple sources with large discrepancy, and only one image is hard to represent all the variations of the data. 
However, when more annotations are available, KAMP can scale much better than CC2D and EGT since our two-stage learning scheme can capture the relation of old and new keypoints and distill them back to the model to improve the old keypoints, which has not been handled in CC2D and EGT. 

\begin{table}[htbp]
\small
  \centering
    \begin{tabular}{l|cccc}
    \toprule
    \textbf{Method/(PCK~\cite{Chen_2024_CVPR} $\uparrow$)} & 1-shot & 5-shot & 10-shot & 50-shot \\
    \midrule
    UKL~\cite{he2022autolink}   & 54.95 & 59.11 & 62.53 & 64.36 \\
    MetaPoint+~\cite{Chen_2024_CVPR}  & 65.71 & 66.87 & 67.26 & 68.98 \\
    \midrule
    KAMP (Ours) & \textbf{70.09} & \textbf{72.23} & \textbf{73.18} & \textbf{76.97} \\
    KAMP (Ours) + \cite{Chen_2024_CVPR} & \textbf{73.49} & \textbf{75.10} & \textbf{77.76} & \textbf{79.18} \\
    \bottomrule
    \end{tabular}%
  \caption{Compare to low-shot methods on Split MPII.}
  \label{comparison to few shot 2D methods}%
  \vspace{-5pt}
\end{table}%
 
As shown in Tab.~\ref{comparison to few shot 2D methods}, we observe that while the CAPE~\cite{xu2022pose,Shi_2023_CVPR,Chen_2024_CVPR} and IKL are two disentangled paradigms, they can complement each other effectively. CAPE relies on pretraining with varied datasets, while IKL focuses on incremental learning, making their training costs incomparable. For example, CAPE methods like MetaPoint+~\cite{Chen_2024_CVPR} require a substantial training cost (4 GPUs, 1.5 days), whereas the `pretrain+incremental' approach of IKL is more efficient (1 GPU, 5 hours).
To demonstrate that CAPE and IKL are orthogonal and complementary, we applied IKL using our KAMP method on MetaPoint+. As shown in Tab.~\ref{comparison to few shot 2D methods}, this combined approach results in significant improvements, even in low-data scenarios, highlighting that IKL can substantially enhance keypoint estimation when applied on a CAPE-pretrained keypoint detector.

The goal of comparing these paradigms for novel keypoint estimation within the same object category is to establish the necessity of IKL. Although CAPE does not involve continual learning and appears more deploy-friendly, its performance plateaus and does not scale effectively with additional data. In contrast, our KAMP achieves higher accuracy with limited data and demonstrates superior scalability with increased data, as evidenced in Tab.~\ref{comparison to few shot 2D methods}.

Lastly, as the first study to explore the IKL paradigm, we focus on scaling up existing keypoint estimators for novel keypoints within the same object category, laying the foundation for future research. While CAPE can perform keypoint estimation on novel object categories with a few support images, its generalization remains limited. 
In the future, we will explore extending the IKL paradigm to novel object categories by applying IKL to keypoint estimators pretrained with CAPE, aiming to combine the strengths of both paradigms to develop a keypoint estimator capable of continually learning new keypoints across diverse object categories without forgetting and with effective scalability.

In summary, the comparison of Tab.~\ref{comparison to few shot medical methods} and~\ref{comparison to few shot 2D methods} show that our proposed IKL paradigm is highly label-efficient for acquiring new keypoints.
This characteristic is advantageous for real-world applications where obtaining sufficient labels is time-consuming, such as in medical analysis~\cite{Cao2023-wf}.

\section{Conclusion}
 We explore learning the newly defined keypoint incrementally without retaining any old data, called Incremental object Keypoint Learning (IKL). We propose a two-stage learning scheme as a novel baseline tailored to the IKL. Extensive experiments show that our method can effectively alleviate the forgetting issue and may even improve the old keypoints' estimation during IKL. Our further analysis reveals that the IKL is label efficient in acquiring the new keypoints, which is promising for real-world applications.

\textbf{Acknowledgements} This work was supported in part by National Science Foundation grant IIS-2007613. This work was also supported by the National Natural Science Foundation of China (62376011).
{
    \small
    \bibliographystyle{ieeenat_fullname}
    \bibliography{main}
}


\clearpage


\setcounter{page}{1}
\onecolumn
\renewcommand{\contentsname}{Contents}
\tableofcontents 
\addtocontents{toc}{\protect\hypersetup{linkcolor=black}}
\appendix

\addtocontents{toc}{\protect\setcounter{tocdepth}{3}}

\section{Guideline of Supplementary}
In this supplementary, in Sec.~\ref{NCO}, we first provide more discussion of the non-co-occurrence~(NCO) problem among Incremental Object Detection~(IOD), Incremental Semantic Segmentation~(ISS), and Incremental Object Keypoint Learning~(IKL) to highlight the novel contribution to resolving the NCO issue in IKL.

in Sec.~\ref{Different Proposes}, we first provide further discussions on the different purposes of our two distillation losses designed for the Stage-II training in our proposed KAMP method, i.e., our Knowledge Association loss $\ell_{K A}$ created by the KA-Net and our Keypoint Spatial-oriented Distillation loss $\ell_{KSD}$ created by the old model $m_{t-1}$. 

In Sec.~\ref{Difference between KA-Net and the Old Model}, we empirically compare the performance of the KA-Net and the old model $m_{t-1}$ on predicting the old keypoints selected in Stage-I and show that they are \textbf{not} similar, as the anatomical prior captured in our KA-Net can further improve the estimations of related old keypoints. This further verifies that the loss $\ell_{K A}$ by KA-Net and the loss $\ell_{KSD}$ by the old model may perform different kinds of knowledge distillation in Stage-II.

As mentioned in our \textbf{main paper Sec.~4.2}, in Sec.~\ref{ablation study on Head-2023} and~\ref{Ablation study on different backbone other than HR-Net}, we provide the ablation study on the Head-2023 dataset and also test our method on other network backbone. In Sec.~\ref{Train KA-Net with Ground-truth Labels}, we provide further ablation study that trains the KA-Net with the ground truth label of the selected old keypoint instead of using the old model $m_{t-1}$ to provide the pseudo-label, where their performances are \textbf{almost the same}, demonstrating that it is \textbf{not} critical to use completely accurate labels to train KA-Net to achieve great results for our method in IKL. 

In Sec.~\ref{complementary property}, we also provide further ablation study on only using the loss $\ell_{K A}$ without the loss $\ell_{KSD}$ in Stage-II training of our method, where we show that it is necessary to use both the loss $\ell_{K A}$ and $\ell_{KSD}$ simultaneously given their different functionalities and complementary property.

Then, we discuss the limitations of our proposed method in Sec.~\ref{limitations}, as mentioned in our \textbf{main paper Sec.~4.3}. In Sec.~\ref{analysis}, we discuss the extreme case when only one new keypoint is considered for constructing the KA-Net in Sec.~\ref{sec: only one keypoint} as mentioned in our \textbf{main paper Sec.~3.2.1}, and more details of the KA-Net. We further provide concrete examples of constructing the auxiliary task in Sec.~\ref{sec: concrete examples}, and provide discussion and ablation study of different Softmax alternatives in Sec.~\ref{sec: different Softmax} and~\ref{sec: ablation study of softmax} respectively.

As mentioned in our \textbf{main paper Sec.~4}, starting from Sec.~\ref{sec: more details and results}, we provide more dataset statistics and experimental details in Sec.~\ref{sec: dataset and experimental details} and include per-step performance and the standard deviation of each dataset in our main paper's Table 1 in Sec.~\ref{sec: std of table 1 and 2}. In Sec.~\ref{sec: more details of evaluation metrics}, we provide more details of our evaluation metric as mentioned in our \textbf{main paper Sec.~4}. In Sec.~\ref{sec: details of low-shot regime}, we provide more details of our adaptation for the low-shot regime as mentioned in our \textbf{main paper Sec.~4.3}. The details of the keypoint group and analysis of the $\alpha$ are included in Sec.~\ref{sec: details of keypoint group} and Sec.~\ref{sec: analysis of alpha} respectively, as mentioned in our \textbf{main paper Sec.~4}. We also report more results over a balanced experimental protocol in Sec.~\ref{sec: more results of balanced setup}, and another challenging experimental protocol when old keypoints are all from the human upper body and new keypoints are all from the human lower body in Split MPII in Sec.~\ref{sec: more results of another setup}. Then we provide more experimental details about the comparison experiments between IKL and other alternative settings (i.e., CC2D~\cite{yao2021one}, EGT~\cite{yin2022one}, UKL~\cite{he2022autolink}, and CAPE~\cite{Shi_2023_CVPR}) in Sec.~\ref{sec: more details about the comparison Experiments} as mention in \textbf{main paper Sec.~4.3}. Finally, we present the per-keypoint transfer metric of Split ATRW after three incremental steps in Sec.~\ref{sec: per-keypoint peformance} as mentioned in our \textbf{main paper Sec.~4.1}, and more visualization results in Sec.~\ref{sec: more visualizations}.

\section{More Discussion of the non-co-occurrence~(NCO) problem among Incremental Object Detection~(IOD), Incremental Semantic Segmentation~(ISS) and Incremental Object Keypoint Learning~(IKL)}
\label{NCO}

While the non-co-occurrence (NCO) problem exists in IOD, ISS, and IKL, the nature and implications of NCO are fundamentally different: (1) In IOD/ISS, NCO affects discrimination between distinct object categories (e.g., cat \& dog). The challenge is primarily about maintaining class boundaries when old classes are not labeled in new data.
(2) In IKL, NCO manifests within the same object category, where new and old keypoints have inherent anatomical and physical relations. The challenge is not class discrimination, but capturing these intrinsic relationships to improve keypoint estimation. 

This unique context of NCO in IKL motivated our novel technical solution: KAMP explicitly models and leverages the relationships between old and new keypoints, improving the estimation of both beyond mere anti-forgetting. This approach is specifically designed for the keypoint estimation context and differs fundamentally from IOD/ISS solutions.

\begin{figure*}[t]
    \centering
    \includegraphics[width=0.85\linewidth]{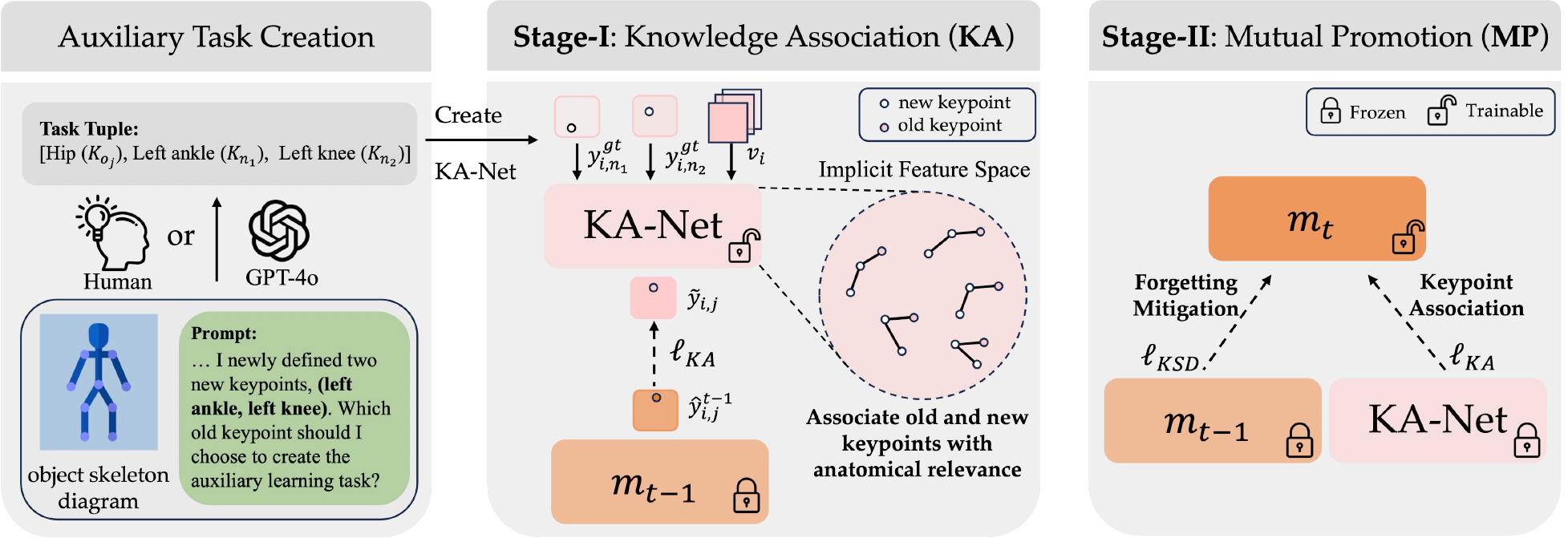}
    \caption{Overview of KAMP using the human body for illustration. In Stage-I, we learn an auxiliary KA-Net to associate the related old and new keypoints based on their local anatomical constraint. In Stage-II, we jointly leverage the old model and the KA-Net as an auxiliary teacher to consolidate all old keypoints' prediction and also learn the new keypoints simultaneously to achieve mutual promotions.} 
    \label{fig:illustration of case 2 network and training_supp}
    \vspace{-15pt}
\end{figure*}

\section{Different Proposes of Knowledge Distillation between $\ell_{K A}$ and $\ell_{KSD}$ in the Stage-II.}
\label{Different Proposes}
As stated in our main paper Sec.~3.2.1, in Stage-II training, the Knowledge Association loss $\ell_{K A}$ is created by the frozen KA-Net, which is learned in Stage-I to capture the implicit anatomical and physical prior between the related old and new keypoint. We leverage $\ell_{K A}$ in Stage-II to distill the \textbf{keypoint association knowledge} to further improve \textbf{the selected old keypoints} in $\mathcal{K}_{KA}$ defined in our main paper Sec.~3.2.1. 

While for the loss $\ell_{KSD}$, as stated in our main paper Sec.~3.2.2, since the loss $\ell_{K A}$ only applies to \textbf{the selected old keypoints} to distill their keypoint association knowledge instead of mitigating the forgetting of all the keypoint, thus the loss $\ell_{KSD}$ is used to distill \textbf{the old model's knowledge} during Stage-II training to consolidate the knowledge for \textbf{all the keypoints} to avoid catastrophic forgetting. 

Therefore, in Stage-II training, for \textbf{the selected old keypoint} (\textcolor{orange}{orange} in Fig.~\ref{fig:illustration of case 2 network and training_supp}), they are supervised by both the loss $\ell_{K A}$ and $\ell_{KSD}$, where the loss $\ell_{K A}$ is for distilling their new knowledge of the keypoint association with the related new keypoints, and the loss $\ell_{KSD}$ is for distilling their previous knowledge from the old model $m_{t-1}$. For other old keypoints, they are all supervised by the loss $\ell_{KSD}$ for knowledge consolidation. 

One may be concerned about two factors: (1) As both the KA-Net and old model can predict \textbf{the selected old keypoints}, it would be interesting to see whether the KA-Net may be similar to the old model $m_{t-1}$ on predicting \textbf{the selected old keypoints} such that the KA-Net may not provide a different distillation effect for the \textbf{the selected old keypoints} compared to the old model during Stage-II. (2) As the KA-Net is trained by the pseudo-label provided by the old model $m_{t-1}$ in Stage-I since we will not label the previously-learned old keypoints during IKL, then given that the prediction from the old model is not as perfect as the ground-truth label, it would be also interesting to see whether the performance of our propose KAMP method may be influenced by the training of the KA-Net. 

To study these two concerns, we first empirically explore the first concern in Sec.~\ref{Difference between KA-Net and the Old Model} by comparing the performance of \textbf{the selected old keypoints} based on the KA-Net and the old model $m_{t-1}$. We show that the KA-Net is \textbf{not} similar to the old model, and the KA-Net can achieve much better performance on \textbf{the selected old keypoints} than the old model. Then in Sec.~\ref{Train KA-Net with Ground-truth Labels}, we train the KA-Net with the ground-truth labels of the old keypoints to see whether it is critical to use very accurate and perfect supervision for training the KA-Net. Our empirical results demonstrate that there is \textbf{almost no difference} between the performance of our KAMP when the KA-Net is either supervised by the pseudo-label from the old model $m_{t-1}$ or by the ground-truth label, which implies that it is \textbf{not} critical to use completely accurate labels to train the KA-Net to achieve great results.

\subsection{Difference between the Performance of the KA-Net and the Old Model on Predicting the Old Keypoints Selected in Stage-I.}
\label{Difference between KA-Net and the Old Model}
As mentioned in Sec.~\ref{Different Proposes}, here we explore the performance difference between the KA-Net and the old model $m_{t-1}$ on predicting \textbf{the selected old keypoints} by using the 5-Step Split MPII protocol. As shown in the Tab.~\ref{compare old model and KA-Net on SOK}, our KA-Net is \textbf{not} similar to the old model on predicting the selected old keypoint, and our KA-Net are all better than the old model on predicting the old keypoints selected in each incremental step. This is because, as stated in our main paper Sec.~3.2, the selected old keypoint predicted by the KA-Net is conditioned on the newly-defined keypoints that are not defined when training the old model. Thus, compared to the prediction from the old model, the prediction of the selected old keypoint from the KA-Net is supplemented with the new anatomical knowledge between the old and new keypoint. Such a physical prior can support the old keypoint prediction by constructing a local constraint to improve the prediction robustness of the old one, and thus, the prediction from the KA-Net for the selected old keypoint is better than the corresponding prediction from the old model.

Therefore, given the difference between the KA-Net and the old model, the KA-Net can further serve as an auxiliary teacher that is different from the old model in Stage-II training to further improve the performance of our KAMP method, which has been verified in our ablation study in our main paper's Sec. 4.2 and Tab. 2.

\begin{table}[htbp]
\small
  \centering
    \begin{tabular}{l|c|c|c|c}
    \toprule
    SOK in each step  & Step-1 & Step-2  & Step-3  & Step-4 \\
    \midrule
    Old Model  $m_{t-1}$ & 64.39 & 68.27 & 77.32 & 95.70 \\
    \midrule
    KA-Net & \textbf{67.36} & \textbf{72.78} & \textbf{83.90} & \textbf{96.32} \\
    \bottomrule
    \end{tabular}%
  \caption{Comparison between the old model $m_{t-1}$ and our KA-Net on predicting \textbf{the Select Old Keypoint} (SOK) in each incremental step in 5-Step Split MPII.}
  \vspace{-15pt}
  \label{compare old model and KA-Net on SOK}%
\end{table}%

\subsection{Ablation study on Head-2023~(\textbf{main paper Sec.~4.2)}}
\label{ablation study on Head-2023}
As mentioned in our main paper Sec.~4.2, here we replicate the ablation study of Tab.~2 in our main paper on Split Head-2023. As shown in Tab.~\ref{tab:ab_Split_Head}, we can achieve the same conclusion stated in our main paper's Sec.~4.2.

\begin{table}[htbp]
\small
  \centering
    \begin{tabular}{l|ccc}
    \toprule
    Method & $\text{A-MRE}_{4}$~$\downarrow$       & $\text{AT}_{4}$     & $\text{MT}_{4}$  \\
    \midrule
    LWF~\cite{li2017learning}   & 4.31  & -1.26 & 0.57 \\
     \midrule
    KAMP~(only $\ell_{KSD}$) & 3.29  & -0.12 & 0.63 \\
    \midrule
    KAMP (Random KA-Net) & 2.93  & 0.08  & 0.73 \\
    \midrule
    KAMP~(Ours)  & \textbf{2.32} & \textbf{0.41} & \textbf{0.84} \\
    \bottomrule
    \end{tabular}%
    \caption{Ablation Study on Split Head-2023}
  \label{tab:ab_Split_Head}%
  \vspace{-15pt}
\end{table}%

\subsection{Ablation study on different backbone other than HR-Net~(\textbf{main paper Sec.~4.2})}
\label{Ablation study on different backbone other than HR-Net}
As stated in our main paper Sec.~3, our KAMP design is general, versatile, and usable with various backbones~\cite{liang2020instance,huang2020dianet,cai2020learning,huang2025generic,Yang_2021_ICCV,li2021tokenpose,xu2022vitpose}. Here, we leverage one of the state-of-the-art methods, i.e., Residual Steps Network~(RSN)~\cite{cai2020learning}, as our backbone, and we can achieve 81.45\% $\text{AAA}_{4}$ on Split MPII, which further outperform 79.93\% obtained by using HRNet backbone in our Table 1 in the main paper. This verifies the generality of our proposed KAMP method.

\subsection{Train KA-Net with Ground-truth Labels.}
\label{Train KA-Net with Ground-truth Labels}
As mentioned in Sec.~\ref{Different Proposes}, here we further explore whether it is critical to use completely accurate supervision to train the KA-Net to achieve a great result for our KAMP method. As shown in Tab.~\ref{tab: Training the KA-Net by GT}, we can observe that even if we use the ground-truth labels of the selected old keypoint to train the KA-Net, the overall average performance ($\text{AAA}_{4}$) of 5-Step Split MPII~(79.97\%) is almost the same as Ours~(79.93\%) that uses the pseudo-label provided by the old model $m_{t-1}$. This shows that it is \textbf{not} so critical to use completely accurate labels to train the KA-Net to achieve great results.

\begin{table}[htbp]
  \centering
    \begin{tabular}{l|ccc}
    \toprule
          & \multicolumn{3}{c}{5-Step Split MPII} \\
    \midrule
    Method & $\text{AAA}_{4}$  & $\text{AT}_{4}$   & $\text{MT}_{4}$\\
    \midrule
    Ours (KA-Net trained by GT) & \textbf{79.97} & 1.70  & \textbf{5.63} \\
    \midrule
    Ours  & 79.93 & \textbf{1.80} & 4.23 \\
    \bottomrule
    \end{tabular}%
    \caption{Training the KA-Net by using the pseudo-label provided by the old model $m_{t-1}$, i.e., Ours, and by using the corresponding ground truth~(GT) labels, i.e., Ours (KA-Net trained by GT).}
    \vspace{-15pt}
  \label{tab: Training the KA-Net by GT}%
\end{table}%

\begin{table}[htbp]
 \small
  \centering
    \begin{tabular}{l|ccc}
    \toprule
    Method & $\text{AAA}_{4}$     & $\text{AT}_{4}$     & $\text{MT}_{4}$  \\
    \midrule
    LWF~\cite{li2017learning}   & 75.75 & -3.86 & 0.41 \\
    \midrule
    KAMP~(only $\ell_{K A}$) & 54.46 & -12.74 & -3.45 \\
    \midrule
    KAMP~(only $\ell_{KSD}$) & 76.93 & -2.24 & 0.65 \\
    \midrule
    KAMP~(Ours)  & \textbf{79.93} & \textbf{1.80} & \textbf{4.23} \\
    \bottomrule
    \end{tabular}%
    \caption{More Ablation Study on 5 Step Split MPII.}
    \vspace{-15pt}
  \label{more ablation on only KA}%
\end{table}%

\subsection{The Necessity of Using $\ell_{K A}$ and $\ell_{KSD}$ Simultaneously.}
\label{complementary property}
As we mentioned in our main paper Sec.~3.2.2, we emphasize that only the loss $\ell_{K A}$ is not enough to mitigate the forgetting of \textbf{all the old keypoints}, since the loss $\ell_{K A}$ is only applied to \textbf{the selected old keypoint} and its functionality is only to distill the keypoint association knowledge to improve the predictions of selected old keypoints. Thus the usage of the loss $\ell_{KSD}$ is necessary as its functionality is to consolidate the knowledge of all the old keypoints based on the old model $m_{t-1}$ to mitigate the forgetting problem. To verify this claim, we extend the ablation study in our main paper Sec.~4.2 and Tab.~2, where we add one more experiment, i.e., only using the $\ell_{K A}$ without the $\ell_{KSD}$ in Stage-II training. The results are shown in Tab.~\ref{more ablation on only KA}, and we can observe that with only $\ell_{K A}$ in Stage-II training, the old keypoints are catastrophically forgotten on 5-Step Split MPII. Only when we use both the loss $\ell_{K A}$ and $\ell_{KSD}$ simultaneously, i.e., Ours, can we achieve the best result. This demonstrates that the different functionalities of the loss $\ell_{K A}$ and $\ell_{KSD}$ make it necessary to use them simultaneously such that we can effectively mitigate the catastrophic forgetting of the old ones and then further improve them. By comparing the alternative of only using the loss $\ell_{KSD}$ in Stage-II and Ours, we can see that the loss $\ell_{K A}$ is complementary to the loss $\ell_{KSD}$ as adding the loss $\ell_{K A}$ to the loss $\ell_{KSD}$ can help us further achieve larger improvement on both the average performance~(i.e., $\text{AAA}_{4}$) and the old keypoints~(i.e., $\text{AT}_{4}$ and $\text{MT}_{4}$). This further implies the complementary property of the loss $\ell_{KSD}$ and $\ell_{K A}$.

\section{Discussions of the Limitations.~(\textbf{main paper Sec.~4.3})}
\label{limitations}
As mentioned in our main paper's Sec.~4.3, since we do not have the ground-truth labels for the old keypoints, thus we leverage the old model's prediction of the old keypoints as the pseudo-label to supervise the KA-Net. However, there would be a concern that the old model's prediction may not be accurate enough to provide supervision as good as the ground-truth label for the old keypoints. 

Our discussions towards this concern are two-fold: (1) in this paper, we actually do not need the KA-Net to be perfect enough to be an auxiliary teacher for the old keypoints. Instead, we hypothesize that even a not strong enough teacher like KA-Net could already benefit the IKL since the KA-Net has different functionality, i.e., it is used to implicitly distill the knowledge of keypoint association into the new model during the IKL. Our empirical study in both the main paper and our supplementary all verified our hypothesis since all the experiments were conducted without concern about whether the KA-Net is strong enough to predict the old keypoints. The empirical result in Sec.~\ref{Train KA-Net with Ground-truth Labels} and Tab.~\ref{tab: Training the KA-Net by GT} further supports that the completely accurate supervision for training the KA-Net is \textbf{not} critical to achieving great results for our method in IKL. (2) In the future, we can further explore leveraging the technique of uncertainty estimation to filter out those uncertain predictions from the old model when training the KA-Net, such that we can prevent the potential low-quality old keypoints' predictions from unstabilizing the training of KA-Net.  

Lastly, regarding our adaptation to the low-shot regime, we employ the same training strategy as outlined in~\cite{yao2021one} to train an auxiliary model in a self-supervised manner. This model is then used to pseudo-label new keypoints when only a few annotations are available in the new data. Consequently, as demonstrated in Tables 3 and 4 of our main paper, the quality of our pseudo-labeling may be constrained by the limitations of~\cite{yao2021one} in extreme 1-shot and/or 5-shot scenarios. We anticipate that these limitations could be overcome with future developments in self-supervised learning. Furthermore, it's important to highlight that our adaptation strategy is primarily introduced to demonstrate the feasibility of our KAMP method in extremely low-shot conditions and to provide a comparative analysis with other low-shot methods. However, in practical applications where high accuracy in keypoint detection is crucial, such as in medical analysis, we generally prefer algorithms that can scale performance with an increase in available labels. In this regard, our KAMP method is advantageous over other alternatives. It not only performs better in low-shot scenarios but also scales more effectively with additional labeled data. For example, labeling 10 to 50 images, which is a manageable task even in medical contexts, can significantly enhance the training of a reliable keypoint detector. As shown in Tables 3 and 4 of our main paper, our KAMP method uniquely scales up with more labels, making it a more favorable option for real-world applications compared to other alternatives.

\section{More Analysis of the Proposed Method.}
\label{analysis}

\subsection{Extreme Case When Only Considering One New keypoint for Constructing the KA-Net.~(\textbf{main paper Sec.~3.2.1})}
\label{sec: only one keypoint}
As mentioned in our main paper Sec.~3.2.1, the auxiliary task construction for the KA-Net can also be created like this: $P(K_{j}^{\text {old}}) = F(P(K_1^{\text {new}}), P(K_{i}^{\text {old}}))$, where when we only have one new keypoint $K_1^{\text {new}}$ that is related to the old keypoint $K_{j}^{\text {old}}$, we can consider other old keypoint $K_{i}^{\text {old}}$ that it is also related to $K_{j}^{\text {old}}$ and $K_1^{\text {new}}$. This shows the generality of our proposed method in that it is still feasible when the extreme case occurs, e.g., only one new keypoint is introduced. To empirically verify the feasibility, we provide a case study using Split MPII and only introduce one new keypoint in each incremental step, as shown in Tab.~\ref{tab:only one keypoint}. We can observe that our method can still achieve a positive average transfer and positive maximal transfer under such an extreme scenario, where we still outperform the competitive baseline, i.e., LWF~\cite{li2017learning}, with a large margin.
\begin{table}[htbp]
  \centering
    \begin{tabular}{l|ccc}
    \toprule
          & \multicolumn{3}{c}{5-Step Split MPII} \\
    \midrule
    Method & $\text{AAA}_{4}$    & $\text{AT}_{4}$   & $\text{MT}_{4}$\\
    \midrule
    LWF~\cite{li2017learning}   & 75.75 & -3.86 & 0.41 \\
    \midrule
    KAMP~(Ours)  & \textbf{79.14} & \textbf{1.41} & \textbf{4.74} \\
    \bottomrule
    \end{tabular}%
    \caption{Results when only one new keypoint is considered for constructing the KA-Net.}
    \vspace{-15pt}
  \label{tab:only one keypoint}%
\end{table}%

\subsection{More Details of the KA-Net}
\label{sec: details of KA-Net}
As described in our main paper's Sec.~3.2.1, we extract the spatial-oriented features for each new keypoints that are the input of the KA-Net. As we want the KA-Net to capture the keypoint association between the related old and new keypoint, thus the KA-Net needs to have the capability of modeling the spatial correlation between those keypoints. Therefore, we use a large convolution kernel to capture the long-range correlation between the old and new keypoints' visual features: for the first two convolutional layers in KA-Net, their kernel size is 15$\times$15, and the padding size is 7. The last convolution layer is for generating the heatmap prediction of the old keypoint, and hence its kernel size is 1$\times$1 and padding size is 0. 

\subsection{Concrete Examples of Task Constructions.}
\label{sec: concrete examples}
\begin{figure*}[t]
    \centering
    \includegraphics[width=1\linewidth]{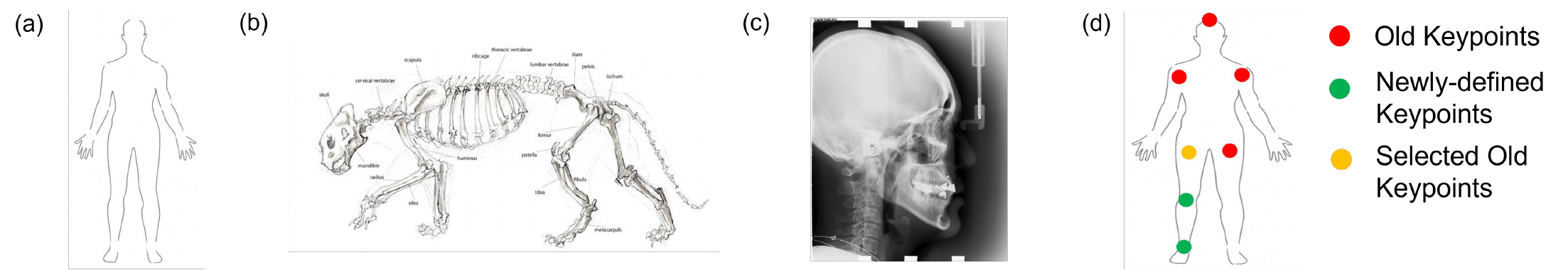}
    \caption{Illustration of the diagram for different objects, e.g., human and tiger body in (a) and (b), and the Cephalometric analysis~\cite{chen2019cephalometric} in (c). All the diagrams of each object category are readily found on the Internet. (d) is an example to demonstrate how we leverage the general diagram, e.g., the human skeleton diagram, to construct the auxiliary prediction task for training the KA-Net in IKL.} 
    \label{fig: diagram}
\end{figure*}

\begin{figure}[ht]
    \centering
    \includegraphics[width=0.5\linewidth]{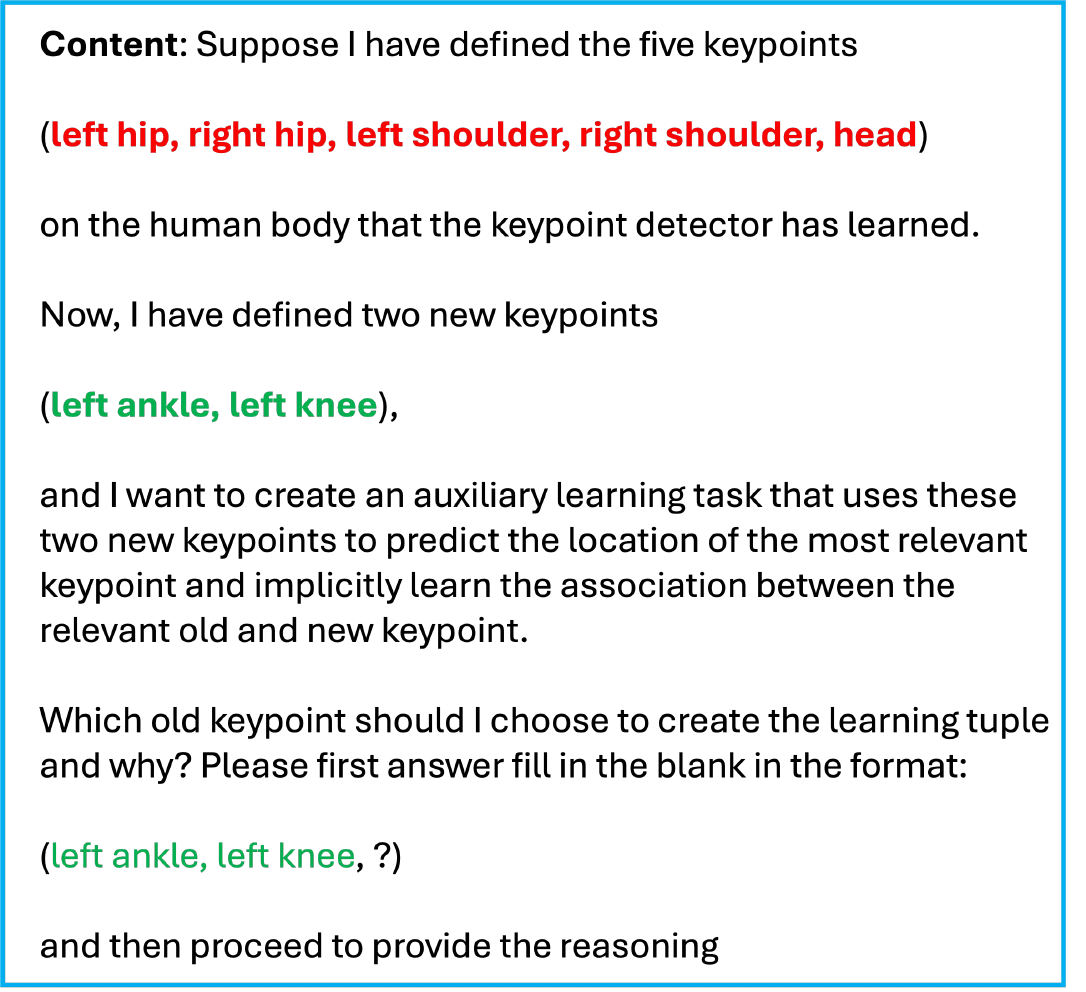}
    \caption{Illustration of the prompting template using the human body as an example to query GPT-4o to create this auxiliary task.} 
    \label{fig: template of GPT-4o}
\end{figure}
In our main paper Sec.~3 and as shown in Fig.~\ref{fig:illustration of case 2 network and training_supp}, when we consider a category for keypoint estimation, it is feasible to create an auxiliary prediction task based on a general object anatomy diagram. These diagrams are readily available on the internet, and a human can interpret them by understanding the semantic meanings of both old and new keypoints. In some medical imaging applications, a doctor may need to clarify the definition of keypoints, but this is a \textbf{one-time} task and is significantly faster than having a doctor label every image. 

Utilizing a general object anatomy diagram for keypoint association eliminates the need to label training images to learn which keypoints should be associated, thus saving substantial annotation and training costs. This approach also provides an interpretable and flexible method for humans to apply physical knowledge of an object category to facilitate incremental learning for the first time. We compare this method with an alternative that involves feeding new data to the old model to obtain pseudo-locations of all old keypoints, and then measuring the relative distances between new and old keypoints to create the auxiliary task. Our method not only reduces the time required from 5.42 minutes to 0.594 seconds but also improves the  $\text{AAA}_{4}$  from 78.82\% to \textbf{79.93\%} for the Split MPII dataset. This improvement is due to the fact that the pseudo-locations predicted by the alternative method may be incorrect, leading to improper creation of the auxiliary task.

To provide concrete examples, Fig.~\ref{fig: diagram} (a), (b), and (c) show that an object anatomy diagram can be easily found online. Since we use this diagram only to identify the proximity of relative locations between old and new keypoints, it can be quite general. Based on this diagram, when learning new keypoints, we first locate both old and new keypoints on the diagram, as illustrated in Figure~\ref{fig: diagram} (d), using their semantic definitions. We then iterate over each old keypoint to find the two closest newly defined keypoints, such as the \textcolor{YellowOrange}{orange} keypoint in Figure~\ref{fig: diagram} (d). Using these three keypoints, i.e., the \textcolor{YellowOrange}{orange} one and the two \textcolor{Green}{green} ones in Figure~\ref{fig: diagram} (d), we construct the auxiliary prediction task for training the KA-Net. 

This task creation can also be automated by GPT-4o as shown in Fig.~\ref{fig:illustration of case 2 network and training_supp}. We only need to create the prompt to provide the name of the old keypoints that the model has learned previously and the name of the newly-defined keypoint. The prompt template is shown in Fig.~\ref{fig: template of GPT-4o}. Empirically, we found that GPT-4o outputs the same task tuple as humans identified for all our experiments, making it a feasible solution to replace the human. The inference cost of prompting the GPT-4o is negligible. 

\subsection{Difference between Softmax Alternatives~(\textbf{main paper Sec.~3.2.2}).}
\label{sec: different Softmax}
\begin{figure*}[t]
    \centering
    \includegraphics[width=1\linewidth]{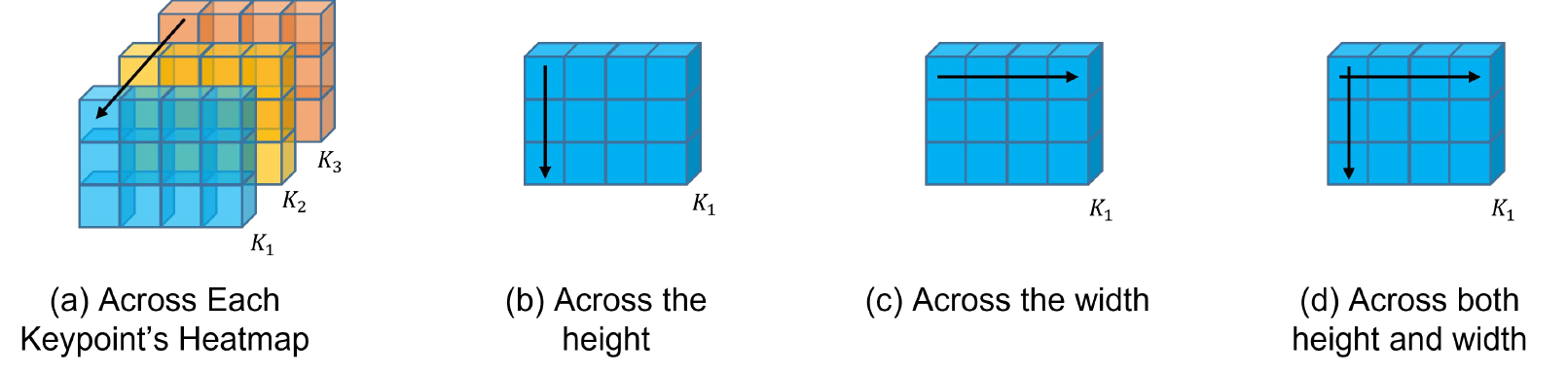}
    \caption{Illustration of different alternatives of Softmax operation.} 
    \label{fig: softmax}
\end{figure*}

As mentioned in our main paper Sec.~3.2.2, We provide the visualization to demonstrate the difference between the Softmax alternatives, as shown in Fig.~\ref{fig: softmax}. Since most existing incremental learning~(IL) literature uses image classification as the default visual task to evaluate the IL methods' performance, thus in methods like LWF~\cite{li2017learning} and its variants~\cite{cha2021cpr}, when they calculate the negative log-likelihood between the old and new model's prediction for the old classes, they all perform the Softmax across difference old classes to obtain the normalized class prediction score. Such an operation in keypoint estimation is equivalent to normalizing each pixel location across each keypoint's heatmap prediction, as shown in Fig.~\ref{fig: softmax} (a), which is a channel-wise normalization. 

However, for the keypoint estimation task, it is always more critical to penalize the spatial-wise correctness for each keypoint individually~\cite{gu2021removing,li2021localization,li20212d,li2021human}. Thus in the present paper, given the task-oriented consideration, we explore the spatial-oriented knowledge distillation, where we perform the softmax over the heatmap spatial dimension, as shown in Fig.~\ref{fig: softmax} (b), (c) and (d), where all these three operations are only normalized each pixel's value over each heatmap itself. Such a principle is also called instance-wise normalization. We will further provide the empirical study in the next section.

\begin{table}[htbp]
  \centering
    \begin{tabular}{l|ccc}
    \toprule
          & \multicolumn{3}{c}{5-Step Split MPII} \\
    \midrule
    Method & $\text{AAA}_{4}$    & $\text{AT}_{4}$   & $\text{MT}_{4}$ \\
    \midrule
    LWF~\cite{li2017learning}   & 75.75 & -3.86 & 0.41 \\
    \midrule
    KAMP~(SM-2D) & 78.54 & 0.52  & 1.78 \\
    KAMP~(SM-Height) & 79.26 & 0.96  & 3.01 \\
    KAMP~(SM-Width) & 79.35 & 1.33  & 3.22 \\
    \midrule
    KAMP~(Ours, Eqn.~5)  & \textbf{79.93} & \textbf{1.80} & \textbf{4.23} \\
    \bottomrule
    \end{tabular}%
    \caption{Ablation study of different Softmax alternatives}
  \label{tab: Softmax}%
\end{table}%

\begin{table}[t]
  \centering
    \begin{tabular}{l|ccc}
    \toprule
          & \multicolumn{3}{c}{5-Step Split MPII} \\
    \midrule
    Method & $\text{AAA}_{4}$    & $\text{AT}_{4}$   & $\text{MT}_{4}$ \\
    \midrule
    Ours ($\alpha$=10) & 78.45 & 0.21  & 0.58 \\
    Ours ($\alpha$=1000) & 79.14 & 0.33  & 2.97 \\
    \midrule
    Ours ($\alpha$=100) & \textbf{79.93} & \textbf{1.80} & \textbf{4.23} \\
    \bottomrule
    \end{tabular}%
    \caption{Analysis of $\alpha$ of the Eqn~(4) in the main paper}
  \label{tab: analysis of alpha}%
\end{table}%
\subsection{More Ablation Study of Different Softmax Alternatives~(\textbf{main paper Sec.~3.2.2})}
\label{sec: ablation study of softmax}

In this section, we explore whether those three spatial-wise Softmax alternatives differ in practice. The results are shown in Tab.~\ref{tab: Softmax}, where SM-2D denotes the Softmax alternative as Fig.\ref{fig: softmax} (d), SM-Height represents the Softmax alternative as Fig.\ref{fig: softmax} (b), and SM-Width represents the Softmax alternative as Fig.\ref{fig: softmax} (c), and Ours which averaging the SM-Width and SM-Height as defined in our main paper Eqn.~(5). 

We can observe that the SM-Width is slightly better than the other two alternatives. By further analysis, we empirically find that for SM-2D, we need to calculate the exponential function for all the pixels over a certain heatmap and summarize them as the denominator of the Softmax operator, leading to total $H$$\times$$W$ terms in the denominator. Such a large denominator makes the value of each pixel tiny after the softmax, making the negative log-likelihood very small. This may further weaken the knowledge distillation effect since its scale will be too small and be less effective than the other two alternatives, i.e., SM-Height and -Width. For SM-Height, we can see that in practice, it is only slightly worse than SM-Width, and we conjecture that the spatial prior may be more readily consolidated via the width dimension. Finally, by combining the SM-Width and SM-Height as defined in our main paper Eqn.~(5), we achieve the best overall performance.

\section{Experimental Details and More Results.}
\label{sec: more details and results}

\subsection{More details of Evaluation Metrics~(\textbf{main paper Sec.~4})}
\label{sec: more details of evaluation metrics}
As mentioned in our main paper, the Probability of Correct Keypoint (PCK)~\cite{yang2021transpose,moskvyak2021semisupervised,wang2022pseudolabeled} is defined as $\text{PCK}$=$\frac{1}{N} \sum_{i=1}^N \mathbf{1}\left(\frac{\left\|y_i-\hat{y}_i\right\|_2}{d} \leq \sigma\right)$, where a predicted keypoint location $\hat{y}_{i}$ is corrected if the normalized distance between $\hat{y}_{i}$ and the ground-truth location $y_{i}$ over the longest side $d$ of the ground-truth bounding box is less than the threshold $\sigma$. For the MPII and ATRW datasets, we use their defaulted $\sigma$ as in \cite{sun2019deep,moskvyak2021semisupervised,Shi_2023_CVPR}.

For mean radial error (MRE)~\cite{yao2021one,yin2022one}, we adhere to the definition in previous studies~\cite{yao2021one,yin2022one}, $\mathrm{MRE}=\frac{1}{N} \sum_i^N \sqrt{\left(x_i-\tilde{x}_i\right)^2+\left(y_i-\tilde{y}_i\right)^2}$, where $(\tilde{x}_i, \tilde{y}_i)$ denotes the predicted location of the keypoint while $(x_{i}, y_{i})$ denotes the ground-truth location.  Since MRE measures the error between the prediction and ground-truth, a smaller value indicates better performance.

The definition of \textbf{A}verage \textbf{T}ransfer is: after step $i$, the average transfer over all previous steps is $\text{AT}_{i}$=$\frac{1}{i-1} \sum_{j=1}^{i-1} (a_{i, j}-a_{j, j})$, where $a_{i, j}$ denotes the average accuracy~(PCK) or error~(MRE) of the keypoints learned at step $j$ after the training of step $i$. We time -1 to the $\text{AT}_{i}$ when $a_{i, j}$ denotes error~(MRE) since MRE is smaller the better. The definition of \textbf{M}aximal \textbf{T}ransfer is: $\text{MT}_i$=$\max _{k \in S_i}\left(R_{k, i}-\gamma_{k}\right)$, which represents the maximal performance change over all old keypoints after step $i$, where $S_{i}$ denotes the set of old keypoints learned before step $i$, $R_{k, i}$ denotes the accuracy~(PCK) or error~(MRE) of the old keypoint $k$ after step $i$, and $\gamma_{k}$ denotes the initial accuracy~(PCL) or error~(MRE) of the old keypoint $k$ when it was first learned in IKL. And we also time MT with -1 when MT measures the MRE.

\subsection{More Dataset Statistics and Experimental Details.~(\textbf{main paper Sec.~4})}
\label{sec: dataset and experimental details}

As mentioned in our main paper, we leverage the Head-2023~\cite{Cao2023-wf}, Chest~\cite{Jaeger2014-zi}, MPII~\cite{andriluka14cvpr}, and ATRW~\cite{li2020atrw} datasets as our main testbed. For the Head-2023 dataset, we randomly chose 20 held-out images as our validation set, and randomly selected 80 held-out images as our test set. The Chest dataset only contains 279 training images after filtering by~\cite{zhu2021you}. Given this small amount of training images, we randomly chose 79 held-out images as our test set without setting a validation set. Instead, we use the same training hyperparameter searched by the validation set of Head-2023 to directly use on the Chest dataset, except for the loss scale $\alpha$ as we use the same $\alpha$ scale as Split ATRW to ensure stable training. For the ATRW dataset, we randomly select 5\% held-out images as our validation set and another 5\% held-out image as our test set for the Split ATRW experiments, given the lack of public ground-truth annotations for the official ATRW test set.

For the Split MPII experiments, we leverage the SGD optimizer, where we reduce the learning rate 10 times at 80 epochs. For Split Head-2023, Split Chest, and Split ATRW experiments, we follow the default optimizer for them, i.e., the Adam~\cite{kingma2014adam}, where $\gamma_{1}=0.99$ and $\gamma_{2}=0$. As mentioned in our main paper, the total number of training epochs is 100 for all the methods. For our proposed method, we leverage 20 epochs for training the KA-Net in Stage-I, and we use the remaining 80 epochs for our Stage-II training, such that we can leverage the same number of epochs for both our method and other compared methods for a fair comparison. 

\subsection{More Details for the adaptation of low-shot regime.~(\textbf{main paper Sec.~4.3})}
\label{sec: details of low-shot regime}
When our method needs to adapt for the low-data regime, we follow the same training strategy as~\cite{yao2021one} to train another auxiliary model in a self-supervised manner as~\cite{yao2021one} in the initial step ($t$=0). Specifically, for a given input image, we randomly select a point on the image and then randomly crop a patch containing that point. The same data augmentations as described in~\cite{yao2021one} are applied to this cropped image patch. The input image and the augmented image patch are then processed through two feature extractors, following the training approach of~\cite{yao2021one}. In the low-shot MPII experiment in Tab.~4 of our main paper, as we conventionally utilize the top-down human pose detector~\cite{sun2019deep}, we first identify each human instance in the image through detection. All detection results for the MPII dataset have been previously provided by top-down pose detectors like~\cite{sun2019deep,cheng2020higherhrnet}. Each detected human instance is then cropped from the image, and these cropped images are used as inputs for training the auxiliary model. 

The training epoch for the auxiliary model is 100 and the same learning rate as in Sec.~4 of our main paper. We use the Adam optimizer with $\gamma_{1}=0.99$ and $\gamma_{2}=0$. Note that for the compared method CC2D~\cite{yao2021one} and EGT~\cite{yin2022one} they also have the same or similar self-supervised pertaining stage in their method to adapt for the low-shot regime for keypoint detection in medical images.  

\subsection{Standard Deviations of Tables 1 in Our Main Paper~(\textbf{main paper Sec.~4}).}
\label{sec: std of table 1 and 2}
As described in our main paper's Sec.~4, we report the per-step results and standard deviations of each dataset in Table 1 of our main paper in Tab.~\ref{Split Head-2023}, Tab.~\ref{Split Chest}, Tab.~\ref{Split MPII} and~\ref{Split ATRW}, respectively, where we observe that our method enjoys relatively more minor deviation and consistently outperform all the comparison methods for all the datasets.

\subsection{Analysis of $\alpha$~(\textbf{main paper Sec.~4})}
\label{sec: analysis of alpha}

As mentioned in our main paper's Sec.~4, in Tab.~\ref{tab: analysis of alpha}, we further provide more analysis of the hyperparameter $\alpha$ in Eqn.~(4) in our main paper. For all our experiments in the main paper and supplementary, we use the held-out validation set to determine the $\alpha$, the same strategy as how we determine the hyperparameters of all the comparison methods. Generally, when the $\alpha$ is large, the knowledge consolidation may dominate the Stage-II training in our proposed method, and thus the acquisition of the new keypoint may be hindered. While if the $\alpha$ is too small, the old keypoint may be forgotten catastrophically in IKL. Therefore, there should exist a proper $\alpha$ that can achieve a good balance between the new keypoint acquisition and avoid the catastrophic forgetting of the old keypoints. Results in Tab.~\ref{tab: analysis of alpha} empirically support our analysis, where we can see that when $\alpha$ increases 10 times from 10 to 1000, the proper $\alpha$ is over 100~(1e2) that can achieve the proper average performance ($\text{AAA}_{4}$).

\subsection{Details of the Keypoint Group.~(\textbf{main paper Sec.~4})}
\label{sec: details of keypoint group}

For the Head-2023 dataset, there are 38 keypoints: Nasal root, Nasal bridge, Outer canthus, Inner canthus, Upper incisal tip, Lower incisal tip, Chin tip, Anterior chin, Inferior chin, Posterior chin, Lower anterior tooth plane, Upper anterior incisal point, Superior protuberance, Inferior protuberance, Lower jaw point, Pre-molar anterior chin, Posterior nostril, Anterior nostril, Beauty midpoint, Center point of the mandibular posterior platform, Upper central incisor tip, Starting point of the mandibular incisor, Small lip spicy point, Red point above the nose, Chin apex, Chin apex, Flat base point, PT point, Bolton point, Upper lip fine point, Lower lip fine point, Alveolar anterior chin point, Alveolar inferior chin point, Chin point, Alveolar chin root point, Chin point, Upper lip external point, Lower lip external point. As mentioned in our main paper Sec.~4, we select the first 19 keypoints as our first group and then splits the rest of the keypoints into 4 groups, where we randomly select two or more keypoints for each incremental step. For the Chest dataset~\cite{Jaeger2014-zi,zhu2021you}, it contains six keypoint as the top, the bottom, and the right boundary point of the right lung and the same three keypoints for the left lung.

For the MPII dataset, there are 16 human body keypoints: right ankle, right knee, right hip, left hip, left knee, left ankle, pelvis, thorax, upper neck, head top, right wrist, right elbow, right shoulder, left shoulder, left elbow and left wrist. We randomly select five keypoints for the initial step, i.e., Step-0, and then we randomly select two or more keypoints for each incremental step. The ATRW dataset has 15 keypoints of Amur Tiger: left ear, right ear, nose, right shoulder, right front paw, left shoulder, left front paw, right hip, right knee, right back paw, left hip, left knee, left back paw, tail, and center. We randomly choose 6 keypoints for the Step-0 of Split ATRW and then randomly choose two or more keypoints for each incremental. For instance, for the qualitative results shown in our main paper and the supplementary, we chose the upper neck, left elbow, right wrist, right knee, and left ankle for the Step-0 training of Split MPII; and we selected nose, tail, right back paw, left back paw, right front paw, and left front paw as the keypoint group introduced in Step-0 for the Split ATRW.

\begin{table}[t]
  \centering
    \begin{tabular}{l|ccc}
    \toprule
          & \multicolumn{3}{c}{Step-1} \\
    \midrule
    Method & $\text{AAA}_{1}$    & $\text{AT}_{1}$   & $\text{MT}_{1}$\\
    \midrule
    EWC~\cite{kirkpatrick2017overcoming}   & 67.13 & -19.69 & 0.14 \\
    RW~\cite{chaudhry2018riemannian}    & 59.12 & -14.97 & -9.62 \\
    MAS~\cite{aljundi2018memory}    & 72.12 & -5.27 & 0.44 \\
    LWF~\cite{li2017learning}   & 77.76 & 0.31  & 1.81 \\
    AFEC~\cite{wang2021afec}  & 68.18 & -4.16 & -0.55 \\
    CPR~\cite{cha2021cpr}   & 77.08 & 0.49  & 1.33 \\
    \midrule
    KAMP~(Ours)  & \textbf{79.17} & \textbf{1.94} & \textbf{3.34} \\
    \bottomrule
    \end{tabular}%
    \caption{Result of the balanced setup}
  \label{Tab: balanced setup}%
  \vspace{-20pt}
\end{table}%

\subsection{More Results of Another Setup: Balanced Number of Old and New Keypoints.}
\label{sec: more results of balanced setup}

For all the experiments before this section, in each incremental step, the number of old keypoints introduced previously is always larger than the number of new keypoints introduced in the current step. Under such a setting, methods that can well preserve the performance of old keypoints will outperform others since the performance of old keypoints may dominate the average accuracy metric, i.e., $\text{AAA}$. 

To provide a more comprehensive view of our method, in this section, we consider a balanced setup where only one incremental step is introduced, and the number of new keypoints is the same as the old ones. In such a balanced setup, we can further see whether our method still has superiority over other methods. As shown in Tab.~\ref{Tab: balanced setup}, we can see that the gap between each comparison method is smaller than the gap we observed in our previous experiments. Our proposed method still achieve the largest average accuracy~($\text{AAA}$) positive average transfer~($\text{AT}$) among all the other methods. This further demonstrates that the superiority of our method is general and consistent over different experimental setups.

\subsection{More Results of Another Setup: Old Keypoints Only for the Upper Body, New Keypoints only for the Lower Body in Split MPII.}
\label{sec: more results of another setup}
To further explore whether our proposed KAMP method can consistently perform well under different setups, here we use the MPII dataset to create a 2-Step protocol where the old keypoints are all from the upper body of the human while the newly-defined keypoints are all from the lower body. Such a scenario can be viewed as a kind of ``extrapolation'' as the keypoints of the lower body are all outside of the upper body. There are not so many physical connections between the upper body and the lower body, and thus the locations of the keypoints of the lower body may not be highly related to the keypoints in the upper body. Therefore such a protocol would be much more challenging than our previous protocols. As shown in Tab.~\ref{tab: upper body then lower body}, compared with the competitive method, CPR~\cite{cha2021cpr}, our proposed method can still achieve positive average transfer and maximal transfer under this challenging protocol and also outperform CPR with a large margin on the average performance. The absolute value of the average transfer and maximal transfer for our method is small, which is expected as explained above. However, our method as a novel baseline for IKL still demonstrates its superiority, and it is also promising for us to explore better methods to further boost the performance in the future.

\begin{table}[htbp]
  \centering
    \begin{tabular}{l|ccc}
    \toprule
          & $\text{AAA}_{1}$  & $\text{AT}_{1}$   & $\text{MT}_{1}$ \\
    \midrule
    CPR~\cite{cha2021cpr}   & 81.48 & -5.33 & -0.42 \\
    \midrule
    Ours  & \textbf{84.53} & \textbf{0.04} & \textbf{0.63} \\
    \bottomrule
    \end{tabular}%
 \caption{Experimental results when we first learn the keypoints all from the upper body of the human and then incrementally learn the new keypoints all from the lower body using the MPII dataset. }
  \label{tab: upper body then lower body}%
\end{table}%

\begin{table*}[t]
  \centering
  \small
  \begin{adjustbox}{width=1.0\textwidth,center}
    \begin{tabular}{l|ccc|ccc|ccc|ccc}
    \toprule
          & \multicolumn{3}{c|}{Step-1} & \multicolumn{3}{c|}{Step-2} & \multicolumn{3}{c|}{Step-3} & \multicolumn{3}{c}{Step-4} \\
    \textbf{Method} & $\text{A-MRE}_{1}$~$\downarrow$ & $\text{AT}_{1}$    & $\text{MT}_{1}$    & $\text{A-MRE}_{2}$~$\downarrow$  & $\text{AT}_{2}$    & $\text{MT}_{2}$    & $\text{A-MRE}_{3}$~$\downarrow$  & $\text{AT}_{3}$    & $\text{MT}_{3}$    & $\text{A-MRE}_{4}$~$\downarrow$  & $\text{AT}_{4}$    & $\text{MT}_{4}$ \\
    \midrule
    EWC~\cite{kirkpatrick2017overcoming}   & 4.12$\pm$0.97  & -1.2$\pm$0.19  & 0.02$\pm$0.14  & 6.36$\pm$1.83  & -2.98$\pm$0.65 & -0.87$\pm$0.38 & 8.86$\pm$2.73  & -5.76$\pm$1.87 & -2.78$\pm$0.99 & 10.97$\pm$2.76 & -6.37$\pm$2.14 & -4.76$\pm$1.53 \\
    RW~\cite{chaudhry2018riemannian}    & 3.67$\pm$0.38  & -0.57$\pm$0.32 & 0.28$\pm$0.07  & 4.25$\pm$0.69  & -1.65$\pm$0.39 & -0.43$\pm$0.37 & 5.75$\pm$0.65  & -2.53$\pm$0.76 & -0.91$\pm$0.47 & 6.49$\pm$1.73 & -4.23$\pm$1.12 & -2.88$\pm$0.85 \\
    MAS~\cite{aljundi2018memory}   & 3.87$\pm$0.10  & -0.76$\pm$0.27 & 0.14$\pm$0.15  & 4.37$\pm$0.53  & -1.12$\pm$0.65 & -0.05$\pm$0.18 & 4.87$\pm$0.93  & -1.59$\pm$0.62 & -0.28$\pm$0.81 & 5.31$\pm$0.54 & -2.15$\pm$0.37 & -1.33$\pm$0.38 \\
    LWF~\cite{li2017learning}   & 3.06$\pm$0.04  & 0.06$\pm$0.07  & 0.27$\pm$0.06  & 3.45$\pm$0.39  & -0.43$\pm$0.24 & 0.21$\pm$0.29  & 3.99$\pm$0.25  & -1.34$\pm$0.41 & -0.39$\pm$0.52 & 4.31$\pm$0.26 & -1.26$\pm$0.20 & 0.57$\pm$0.25 \\
    AFEC~\cite{wang2021afec}  & 3.12$\pm$0.05  & -0.04$\pm$0.05 & 0.42$\pm$0.12  & 3.96$\pm$0.65  & -0.84$\pm$0.16 & 0.05$\pm$0.38  & 4.94$\pm$1.02  & -1.65$\pm$0.87 & -0.81$\pm$0.64 & 5.77$\pm$0.94 & -3.45$\pm$0.69 & -1.46$\pm$0.75 \\
    CPR~\cite{cha2021cpr}   & 2.96$\pm$0.02  & 0.29$\pm$0.03  & 0.72$\pm$0.06  & 3.18$\pm$0.12  & -0.06$\pm$0.28 & 0.33$\pm$0.11  & 3.46$\pm$0.16  & -0.76$\pm$0.12 & 0.63$\pm$0.08  & 3.71$\pm$0.41 & -1.18$\pm$0.30 & 0.16$\pm$0.13 \\
    
    SFD~\cite{douillard2021plop} & 3.45$\pm$0.03  & 0.12$\pm$0.11  &  0.02$\pm$0.06  & 3.59$\pm$0.11  & 0.01$\pm$0.05 & 0.13$\pm$0.02  & 4.43$\pm$0.07  & -0.18$\pm$0.06 & 0.19$\pm$0.02  & 4.76$\pm$0.11 & -0.43$\pm$0.07 & 0.02$\pm$0.04 \\
    WF~\cite{xiao2023endpoints} & 3.37$\pm$0.02  & 0.14$\pm$0.03  & 0.35$\pm$0.11  & 3.50$\pm$0.13  & 0.03$\pm$0.02 & 0.19$\pm$0.03  & 4.29$\pm$0.10  & -0.01$\pm$0.11 & 0.21$\pm$0.16  & 4.58$\pm$0.23 & 0.03$\pm$0.09 & 0.11$\pm$0.03 \\
    GBD~\cite{dong2023heterogeneous} & 3.28$\pm$0.04  & 0.21$\pm$0.10  & 0.43$\pm$0.07  & 3.41$\pm$0.09  & 0.02$\pm$0.05 & 0.10$\pm$0.03  & 4.18$\pm$0.08  & 0.02$\pm$0.03 & 0.27$\pm$0.08  & 4.34$\pm$0.17 & 0.12$\pm$0.08 & 0.47$\pm$0.02 \\
    \midrule
    KAMP (Ours)  & \textbf{2.13$\pm$0.04} & \textbf{0.63$\pm$0.08} & \textbf{0.92$\pm$0.12} & \textbf{2.25$\pm$0.07} & \textbf{0.36$\pm$0.06} & \textbf{0.72$\pm$0.19} & \textbf{2.29$\pm$0.03} & \textbf{0.33$\pm$0.02} & \textbf{0.65$\pm$0.07} & \textbf{2.32$\pm$0.09} & \textbf{0.41$\pm$0.03} & \textbf{0.84$\pm$0.09} \\
    \bottomrule
    \end{tabular}%
    \end{adjustbox}
    \caption{Results on Split Head-2023 after 5 Step IKL, starting from the same Step-0 trained model. A-MRE: smaller the better}
  \label{Split Head-2023}%
\end{table*}%

\begin{table}[h]
  \centering
   \footnotesize
    \begin{tabular}{l|ccc}
    \toprule
          & \multicolumn{3}{c}{Step-1} \\
    \textbf{Method} & $\text{A-MRE}_{1}$~$\downarrow$  & $\text{AT}_{1}$ & $\text{MT}_{1}$ \\
    \midrule
    EWC~\cite{kirkpatrick2017overcoming}   & 13.28$\pm$2.31 & -8.23$\pm$1.87 & -3.67$\pm$1.21 \\
    RW~\cite{chaudhry2018riemannian}    & 9.48$\pm$1.53  & -7.12$\pm$1.29 & -4.15$\pm$0.76 \\
    MAS~\cite{aljundi2018memory}   & 7.36$\pm$1.02  & -1.86$\pm$0.83 & -0.19$\pm$0.64 \\
    LWF~\cite{li2017learning}   & 6.35$\pm$0.09  & -1.34$\pm$0.12 & 0.18$\pm$0.07 \\
    AFEC~\cite{wang2021afec}  & 8.04$\pm$0.28  & -2.67$\pm$0.31 & 0.15$\pm$0.0.10 \\
    CPR~\cite{cha2021cpr}   & 6.17$\pm$0.06  & -0.87$\pm$0.03 & 0.29$\pm$0.05 \\
    SFD~\cite{douillard2021plop}  & 7.68$\pm$0.03  & -0.54$\pm$0.04 & 0.13$\pm$0.01 \\
    WF~\cite{xiao2023endpoints}     & 7.31$\pm$0.03  & -0.31$\pm$0.02 & 0.16$\pm$0.04 \\
    GBD~\cite{dong2023heterogeneous}   & 6.42$\pm$0.02  & 0.06$\pm$0.07 & 0.21$\pm$0.03 \\
    \midrule
    KAMP (Ours)  & \textbf{5.67$\pm$0.08} & \textbf{0.29$\pm$0.03} & \textbf{0.62$\pm$0.09} \\
    \bottomrule
    \end{tabular}%
    \caption{Results on Split Chest after 2 Step IKL, starting from the same Step-0 trained model. A-MRE: smaller the better}
  \label{Split Chest}%
\end{table}%

\begin{table*}[t]
  \centering
  \small
  \begin{adjustbox}{width=1.0\textwidth,center}
   \begin{tabular}{l|ccc|ccc|ccc|ccc}
    \toprule
          & \multicolumn{3}{c|}{Step-1} & \multicolumn{3}{c|}{Step-2} & \multicolumn{3}{c|}{Step-3} & \multicolumn{3}{c}{Step-4} \\
    \midrule
    Method & $\text{AAA}_{1}$    & $\text{AT}_{1}$    & $\text{MT}_{1}$    & $\text{AAA}_{2}$    & $\text{AT}_{2}$     & $\text{MT}_{2}$   & $\text{AAA}_{3}$     & $\text{AT}_{3}$     & $\text{MT}_{3}$    & $\text{AAA}_{4}$   & $\text{AT}_{4}$    & $\text{MT}_{4}$ \\
    \midrule
    EWC~\cite{kirkpatrick2017overcoming}   & 52.42$\pm$4.68 & -26.18$\pm$6.6 & -0.95$\pm$ 0.01 & 32.37$\pm$ 5.8 & -50.08$\pm$ 6.61 & -25.53$\pm$ 11.27 & 24.67$\pm$1.75 & -57.59$\pm$ 2.82 & -35.56$\pm$ 12.02 & 38.64$\pm$ 0.40 & -51.84$\pm$0.95 & -12.21$\pm$ 2.69 \\
    RW~\cite{chaudhry2018riemannian}    & 58.85$\pm$0.79 & -3.19$\pm$2.91 & -0.41$\pm$0.007 & 51.12$\pm$1.78 & -6.88$\pm$0.79 & -0.66$\pm$0.14 & 43.44$\pm$0.05 & -8.20$\pm$6.36 & -0.89$\pm$4.46 & 38.47$\pm$2.65 & -18.83$\pm$0.51 & -7.13$\pm$3.72 \\
    MAS~\cite{aljundi2018memory}   & 65.35$\pm$6.14 & -6.97$\pm$7.06 & -0.24$\pm$0.10 & 59.99$\pm$2.00 & -13.85$\pm$2.59 & -0.10$\pm$0.07 & 63.29$\pm$2.89 & -9.91$\pm$3.7 & 0.00$\pm$0.17  & 67.03$\pm$1.65 & -7.56$\pm$0.7 & 0.34$\pm$0.20 \\
    LWF~\cite{li2017learning}   & 71.23$\pm$0.46 & -0.49$\pm$0.35 & 1.41$\pm$0.16  & 73.75$\pm$0.65 & -1.03$\pm$0.60 & 0.28$\pm$0.17  & 74.69$\pm$0.56 & -1.63$\pm$0.38 & 0.68$\pm$0.11  & 75.75$\pm$0.51 & -3.86$\pm$1.76 & 0.41$\pm$0.31 \\
    AFEC~\cite{wang2021afec}  & 63.54$\pm$0.79 & -4.78$\pm$0.44 & -1.50$\pm$0.41 & 52.98$\pm$3.2 & -17.26$\pm$2.37 & -7.33$\pm$1.87 & 47.25$\pm$8.18 & -18.30$\pm$11.32 & -6.17$\pm$8.77 & 37.24$\pm$0.05 & -22.85$\pm$1.06 & -15.42$\pm$0.07 \\
    CPR~\cite{cha2021cpr}   & 71.67$\pm$0.11 & 0.68$\pm$0.04& 2.17$\pm$0.08  & 72.86$\pm$0.42 & -2.42$\pm$0.30 & 0.61$\pm$0.16  & 73.95$\pm$0.36 & -2.07$\pm$0.55 & 1.19$\pm$0.30  & 75.52$\pm$0.67 & -3.24$\pm$1.00 & 0.75$\pm$0.62 \\
    
    SFD~\cite{douillard2021plop}   & 68.79$\pm$0.16 & 0.07$\pm$0.01& 0.53$\pm$0.12  & 70.09$\pm$0.35 & -0.98$\pm$0.33 & 0.71$\pm$0.12  & 71.15$\pm$0.38 & -0.41$\pm$0.28 & 0.85$\pm$0.05  & 71.49$\pm$0.82 & -0.93$\pm$0.52 & 0.21$\pm$0.08 \\
    WF~\cite{xiao2023endpoints}  & 69.92$\pm$0.19 & 0.08$\pm$0.13 & 0.76$\pm$0.06 & 70.94$\pm$0.29 & -1.31$\pm$0.26 & 0.58$\pm$0.18  & 72.24$\pm$0.36 & -1.65$\pm$0.07 & 1.54$\pm$0.14  & 72.87$\pm$0.39 & -0.46$\pm$0.04 & 0.38$\pm$0.14 \\
    GBD~\cite{dong2023heterogeneous}  & 71.94$\pm$0.06 & 0.18$\pm$0.06 & 1.76$\pm$0.09  & 72.69$\pm$0.26 & -1.11$\pm$0.17 & 0.69$\pm$0.05  & 74.74$\pm$0.67 & -0.10$\pm$0.38 & 0.64$\pm$0.18  & 75.62$\pm$0.18 & -0.18$\pm$0.06 & 0.35$\pm$0.19 \\
    \midrule
    KAMP~(Ours)  & \textbf{73.54$\pm$0.34} & \textbf{2.23$\pm$0.37} & \textbf{3.98$\pm$0.25} & \textbf{76.55$\pm$0.11} & \textbf{2.30$\pm$0.04} & \textbf{3.68$\pm$0.02} & \textbf{78.06$\pm$0.49} & \textbf{2.24$\pm$0.35} & \textbf{4.12$\pm$0.33} & \textbf{79.93$\pm$0.12} & \textbf{1.80$\pm$0.11} & \textbf{4.23$\pm$0.31} \\
    \bottomrule
    \end{tabular}%
    \end{adjustbox}
    \caption{Results on Split MPII after 5 Step IKL, starting from the same Step-0 trained model.}
  \label{Split MPII}
\end{table*}%

\begin{table*}[t]
  \centering
  \begin{adjustbox}{width=1.0\textwidth,center}
    \begin{tabular}{l|ccc|ccc|ccc}
    \toprule
          & \multicolumn{3}{c|}{Step-1} & \multicolumn{3}{c|}{Step-2} & \multicolumn{3}{c}{Step-3} \\
    \midrule
    Method & $\text{AAA}_{1}$   & $\text{AT}_{1}$     & $\text{MT}_{1}$   & $\text{AAA}_{2}$    & $\text{AT}_{2}$     & $\text{MT}_{2}$    & $\text{AAA}_{3}$    & $\text{AT}_{3}$    & $\text{MT}_{3}$ \\
    \midrule
    EWC~\cite{kirkpatrick2017overcoming}    & 40.33$\pm$4.06 & -92.58$\pm$8.81 & -81.71$\pm$3.76 & 25.19$\pm$2.63 & -62.88$\pm$6.57 & -17.28$\pm$6.75 & 14.38$\pm$1.01 & -59.75$\pm$8.13 & -2.08$\pm$1.6 \\
    RW~\cite{chaudhry2018riemannian}    & 82.90$\pm$0.32  & -1.09$\pm$1.14 & -0.56$\pm$0.61 & 84.14$\pm$1.63 & -8.10$\pm$2.95 & 0.00$\pm$1.26  & 84.15$\pm$0.43 & -10.87$\pm$2.61 & 0.00$\pm$1.12 \\
    MAS~\cite{aljundi2018memory}   & 89.54$\pm$0.4 & -7.26$\pm$0.45 & -0.56$\pm$1.12 & 88.26$\pm$0.06 & -5.80$\pm$0.22 & -0.62$\pm$1.18 & 85.68$\pm$1.43 & -5.80$\pm$4.31 & -1.13$\pm$0.57 \\
    LWF~\cite{li2017learning}   & 90.45$\pm$0.19 & -6.18$\pm$1.06 & -4.52$\pm$1.69 & 89.15$\pm$0.61 & -4.99$\pm$0.63 & 2.47$\pm$0.74  & 87.31$\pm$1.05 & -5.10$\pm$2.83 & -0.64$\pm$1.21 \\
    AFEC~\cite{wang2021afec}  & 61.57$\pm$1.52 & -30.46$\pm$0.05 & -10.11$\pm$0.69 & 45.70$\pm$0.52  & -35.75$\pm$1.16 & -9.26$\pm$0.98 &  33.03$\pm$0.75 & -40.25$\pm$0.88   & -8.02$\pm$0.15 \\
    CPR~\cite{cha2021cpr}   & 90.86$\pm$0.36 & -3.24$\pm$0.09 & 0.00$\pm$0.69  & 90.43$\pm$0.46 & -2.22$\pm$1.39 & 1.85$\pm$0.78  & 89.34$\pm$0.73 & -2.76$\pm$0.75 & 4.49$\pm$0.12 \\
    
    SFD~\cite{douillard2021plop}   & 88.98$\pm$0.36 & -2.19$\pm$0.11 & -0.42$\pm$0.69  & 87.54$\pm$0.47 & -1.88$\pm$1.09 & 1.18$\pm$0.75  & 86.11$\pm$0.81 & -1.13$\pm$0.21 & 0.41$\pm$0.98 \\
    WF~\cite{xiao2023endpoints}   & 90.16$\pm$0.24 & -2.08$\pm$0.02 & -0.35$\pm$0.52  & 88.63$\pm$0.38 & -1.76$\pm$0.96 & 1.77$\pm$0.68  & 86.69$\pm$0.87 & -0.97$\pm$0.29 & 0.62$\pm$0.54 \\
    GBD~\cite{dong2023heterogeneous}   & 90.86$\pm$0.36 & -3.24$\pm$0.09 & 0.00$\pm$0.69  & 89.03$\pm$0.18 & -1.23$\pm$0.67 & 1.84$\pm$0.98  & 87.42$\pm$0.77 & -0.89$\pm$0.30 & 0.65$\pm$0.25 \\
    \midrule
    KAMP~(Ours)  & \textbf{93.21$\pm$0.76} & \textbf{-0.86$\pm$0.27} & \textbf{0.56$\pm$0.32} & \textbf{93.63$\pm$0.24} & \textbf{-0.34$\pm$0.21} & \textbf{3.08$\pm$0.61} & \textbf{93.16$\pm$0.34} & \textbf{-0.84$\pm$0.28} & \textbf{5.13$\pm$0.47} \\
    \bottomrule
    \end{tabular}%
    \end{adjustbox}
    \caption{Results on Split ATRW after 4 Step IKL, starting from the same Step-0 trained model.}
    \label{Split ATRW}
\end{table*}%

\subsection{More Experimental Details about the Low-shot Experiments between our IKL and other Alternative Methods.~(\textbf{main paper line 580-581})}
\label{sec: more details about the comparison Experiments}
Here we provide more experimental details about the low-shot experiments between our proposed IKL setting and other alternative methods of our Sec.~4.3 in our main paper. For the experiment on Head-2023, since both CC2D~\cite{yao2021one} and EGT~\cite{yin2022one} have not trained on Head-2023, thus we use the official implementation of them to run the experiment on Head-2023 to get the results. All the training details are the same as their original paper~\cite{yao2021one,yin2022one}.

For the unsupervised keypoint learning~(UKL)~\cite{he2022autolink} in Split MPII, we choose the SOTA UKL method~\cite{he2022autolink} to perform our experiments. We follow the same experimental details in \cite{he2022autolink} to conduct the unsupervised pertaining on the MPII datasets, where we pre-define the model to output 32 keypoints without assigning any semantic meaning for each one. Then after the unsupervised pretraining on each dataset, we follow the standard practice in UKL~\cite{zhang2018unsupervised,sanchez2019object,mallis2020unsupervised,he2022autolink} that we freeze the unsupervised pretrained model and then learn a linear transformation between the pre-defined keypoints and each newly-defined keypoint introduced in each incremental step. 

For the category-agnostic pose estimation~(CAPE)~\cite{Chen_2024_CVPR} in Split MPII, we leverage its SOTA method, i.e., MetaPoint+~\cite{Chen_2024_CVPR}, to conduct our experiments. We also follow the same experimental details in \cite{Chen_2024_CVPR}, where we treat the keypoint categories related to humans~(e.g., human body, face, and hand) as the unseen category to avoid information leakage when performing the pertaining in CAPE. This is similar to the cross super-category experiments in Sec.~4.3 in \cite{xu2022pose}. The difference is that in our experiments, we need to perform the testing on each keypoint group incrementally.

\begin{table*}[t]
  \centering
    \begin{tabular}{l|c|c|c|c|c|c|c|c|c|c|c|c|c}
    \toprule
          & \textbf{KP1} & \textbf{KP2} & \textbf{KP3} & \textbf{KP4} & \textbf{KP5} & \textbf{KP6} & \textbf{KP7} & \textbf{KP8} & \textbf{KP9} & \textbf{KP10} & \textbf{KP11} & \textbf{KP12} & \textbf{KP13}\\
    \midrule
    Transfer & 1.13  & -2.25 & -2.86 & -14.62 & -1.71 & 1.13  & 2.92  & -0.58 & 5.13  & -3.09 & 2.47  & -1.41 & 0.00 \\
    \bottomrule
    \end{tabular}%
     \caption{Per-keypoint performance transfer in Split ATRW after 3 incremental steps. $\text{KP}$ is the abbreviation of keypoint.}
  \label{tab: per-keypoint}%
\end{table*}%

\subsection{Per-keypoint Performance of Our Method under the ATRW.~(\textbf{main paper Sec.~4.1})}
\label{sec: per-keypoint peformance}

As mentioned in our main paper's Sec.~4.1, here we report the per-keypoint's performance of knowledge transfer after three incremental steps of Split ATRW. As shown in Tab.~\ref{tab: per-keypoint}, 6 over 13 old keypoints have non-negative transfer after three incremental steps. This further verifies our conjecture that there is much positive transfer occurring for many old keypoints to \textbf{offset} the forgetting of other old keypoints  such that our method can achieve a very small negative average transfer.

\subsection{More Visualization Results.~(\textbf{main paper Fig.~4})}
\label{sec: more visualizations}

\begin{figure*}[t]
	\centering
	\includegraphics[width=1\linewidth]{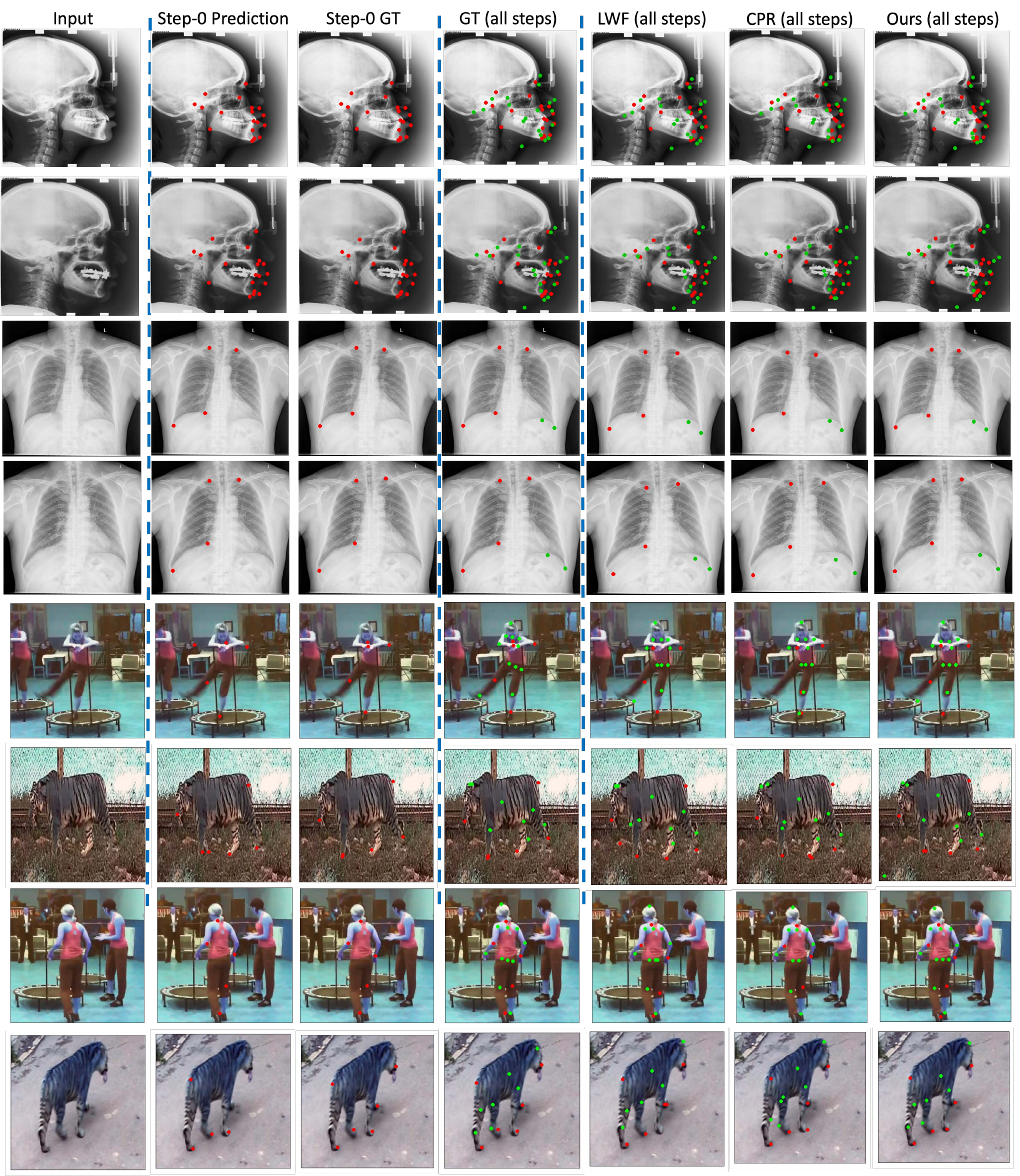}
	\caption{Visualization results on Split Head-2023, Split Chest, Split MPII and Split ATRW. All the methods start from the same Step-0 model, whose prediction is shown in the second column. GT denotes ground truth. The red circles denote the keypoints learned in Step 0, while the green circles denote all the new keypoints learned in later steps. We observe that after the IKL, the compared methods~(LWF and CPR) may acquire the new keypoints as ours, but they have obvious miss-detection and wrong estimation~(e.g., out of the body). While our method can consistently associate the new and old keypoints and achieve structurally accurate keypoint predictions.} 
	\label{fig: qualitative result_supp}
\end{figure*}

As mentioned in our main paper's Fig.~4, we include more visualization results in Fig.~\ref{fig: qualitative result_supp}. Again, our method can achieve more structurally correct keypoint prediction and less miss-detection error than other comparison methods.

\end{document}